\documentclass[10pt,twocolumn,letterpaper]{article}

\usepackage[pagenumbers]{wacv} 

\usepackage{graphicx}
\usepackage{amsmath}
\usepackage{amssymb}
\usepackage{cancel}
\usepackage{pifont}
\usepackage{booktabs}
\usepackage[dvipsnames]{xcolor}
\usepackage{xspace}
\usepackage{textcomp}
\usepackage{array}
\usepackage[skins]{tcolorbox}
\usepackage{float}
\usepackage{newfloat}
\usepackage{afterpage}
\usepackage{numprint}
\usepackage{microtype}
\usepackage[accsupp]{axessibility}  

\usepackage[pagebackref,breaklinks,colorlinks]{hyperref}

\allowdisplaybreaks

\newfloat{promptfloat}{!t}{pop}
\newtcolorbox[auto counter]{prompt}[2]{label={prompt:#1}, title={Prompt \thetcbcounter: #2}, fonttitle=\footnotesize\sffamily, fontupper=\scriptsize\sffamily, fontlower=\scriptsize\sffamily, left=1mm, right=1mm, top=1mm, bottom=1mm, middle=1mm}
\makeatletter
\newcommand\tcb@cnt@promptautorefname{Prompt}
\makeatother
\newcommand{\psystem}[1]{\textcolor{Brown}{\textbf{SYSTEM:}} #1\xspace}
\newcommand{\pprompt}[1]{\textcolor{Green}{\textbf{PROMPT:}} #1\xspace}
\newcommand{\presponse}[1]{\textcolor{Orchid}{\textbf{RESPONSE:}} #1\xspace}

\newcommand{\degree}{\ensuremath{^\circ}\xspace}
\newcommand{\cmark}{\ding{51}}
\newcommand{\xmark}{\ding{55}}
\newcommand{\red}[1]{\textcolor{red}{#1}}
\newcommand{\green}[1]{\textcolor{Green}{#1}}

\usepackage[capitalize,noabbrev,nameinlink]{cleveref}
\crefname{section}{Sec.}{Secs.}
\Crefname{section}{Sec.}{Secs.}
\crefname{appendix}{App.}{Apps.}
\Crefname{appendix}{App.}{Apps.}
\crefname{figure}{Fig.}{Figs.}
\Crefname{figure}{Fig.}{Figs.}
\crefname{table}{Tab.}{Tabs.}
\Crefname{table}{Tab.}{Tabs.}
\crefname{equation}{Eq.}{Eqs.}
\Crefname{equation}{Eq.}{Eqs.}

\def\wacvPaperID{333}
\def\confName{WACV}
\def\confYear{2025}

\begin{document}

\title{Unconstrained Open Vocabulary Image Classification:\\Zero-Shot Transfer from Text to Image via CLIP Inversion}
\author{Philipp Allgeuer \and Kyra Ahrens\\University of Hamburg\\\makebox[2cm][c]{\tt\small\{philipp.allgeuer, kyra.ahrens, stefan.wermter\}@uni-hamburg.de} \and Stefan Wermter}
\maketitle

\begin{abstract}
We introduce NOVIC, an innovative real-time uNconstrained Open Vocabulary Image Classifier that uses an autoregressive transformer to generatively output classification labels as language. Leveraging the extensive knowledge of CLIP models, NOVIC harnesses the embedding space to enable zero-shot transfer from pure text to images. Traditional CLIP models, despite their ability for open vocabulary classification, require an exhaustive prompt of potential class labels, restricting their application to images of known content or context. To address this, we propose an ``object decoder'' model that is trained on a large-scale 92M-target dataset of templated object noun sets and LLM-generated captions to always output the object noun in question. This effectively inverts the CLIP text encoder and allows textual object labels from essentially the entire English language to be generated directly from image-derived embedding vectors, without requiring any a priori knowledge of the potential content of an image, and without any label biases. The trained decoders are tested on a mix of manually and web-curated datasets, as well as standard image classification benchmarks, and achieve fine-grained prompt-free prediction scores of up to 87.5\%, a strong result considering the model must work for any conceivable image and without any contextual clues.\footnote{The authors gratefully acknowledge support from the DFG (TRR 169 -- CML) and European Commission (TRAIL).}
\end{abstract}
\vspace*{-1ex}

\section{Introduction}
\label{sec:introduction}

\begin{figure}[t]
    \parbox{\linewidth}{\centering\includegraphics[width=\linewidth]{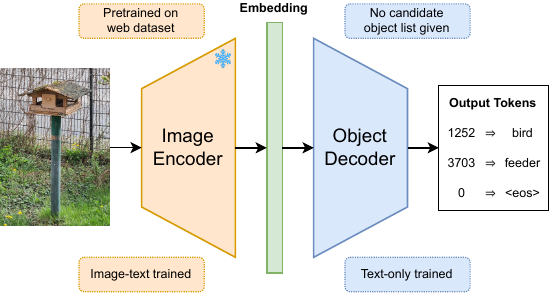}}
    \caption{\textbf{Proposed open vocabulary image classifier.} The classifier processes each input image by encoding it into a CLIP embedding vector, which is then decoded into a sequence of tokens representing the object class. The object decoder is purely generative, producing free-form text without relying on a predefined list of candidate objects. This method is trained solely on textual data and is capable of zero-shot transfer to image data during inference.}
    \label{fig:teaser}
    \vspace*{-1ex}
\end{figure}

Imagine a robot navigating through an environment, encountering a small cup, and an operator asking, \emph{``Can you tell me what that is?''} Ideally, the robot would recognize the object with the objectives to: a) answer \emph{cup} directly, and b) provide an answer immediately in real-time, even if there are large numbers of objects to consider at high frame rates. Contrary to the first objective, Vision-Language Models (VLM) like CLIP \cite{radford_learning_2021} require the robot to be prompted with a comprehensive textual list of object class candidates to perform zero-shot classification. On the other hand, equipping the robot with a high-performing multimodal Large Language Model (LLM) is computationally expensive, requires specialized remote servers for inference, and does not allow for the real-time responsiveness needed for video frame rates, thereby failing to meet the second objective. The core aim of this work is to develop models capable of unconstrained open vocabulary image classification. The models should generate \emph{object nouns} directly as free-form text in real time, without the need for prompts or predefined label candidates.

Current approaches to open vocabulary learning \cite{wu_towards_2024} with VLMs for visual reasoning tasks (\eg Tag2Text \cite{huang_tag2text_2024} and RAM \cite{zhang_recognize_2024} for image tagging) are in general limited by the need for candidate lists, which, with at most a few thousand entries, are not comprehensive enough to encompass all potential object classes. Although some parallel works propose strategies to utilize VLMs for generative tasks \cite{li_decap_2023,gu_close_2023}, these are still ultimately restricted in diversity by the class candidates of labeled training datasets, leaving truly unconstrained prompt-free image classification unresolved. To address this gap in current research, we propose NOVIC, a novel \emph{uNconstrained Open Vocabulary Image Classifier} (cf.~\Cref{fig:teaser}). The classifier can directly label any given image with an English object noun without requiring any category prompts. Thanks to the use of synthetic large-scale text-only datasets, very efficient and scalable training of the so-called \emph{object decoder} transformer is achieved by training purely on CLIP text embeddings. The decoder effectively learns to invert a CLIP text encoder, mapping embeddings back to object nouns via autoregressive token sequence generation. By leveraging the shared multimodal representations of pretrained VLMs, NOVIC can perform zero-shot image classification without ever having seen images during training.

To train the object decoder, we introduce an \emph{object noun dictionary}, and use it to build several synthetically and LLM-generated caption-object datasets. These datasets are all created through automated pipelines, allowing for a high degree of scalability, especially compared to image-based training. Inspired by prior works \cite{nukrai_text-only_2022,gu_close_2023}, noise augmentation is used on CLIP embeddings of the captions during training to bridge the large modality gap between image and text embeddings \cite{liang_mind_2022}. The resulting object decoder can be evaluated on classification datasets like ImageNet-1K \cite{imagenet1k}, but this does not fully capture its prompt-free open vocabulary performance. To address this, we curate three open vocabulary image datasets from original and web sources, and annotate them using human and LLM knowledge.

In summary, our main contributions\footnote{All code \cite{github_novic,github_object_noun_dict} (GitHub) and datasets \cite{corpora_novic,corpora_ovic_datasets} are publicly available.} are as follows: (i) we propose a novel open vocabulary object decoder model that is trained on text only and can zero-shot classify arbitrary images in real-time without any candidates or prompts, (ii) we develop an automated pipeline to construct a comprehensive English object noun dictionary and use a multiset prompting scheme combined with an LLM to construct large-scale synthetic caption-object datasets, (iii) we curate three new image sets intended for evaluation of open vocabulary classification and provide human and multimodal LLM annotations, and (iv) we show that our method scales in performance with the underlying CLIP model, and is able to provide accurate yet very fine-grained classifications while achieving strong in-the-wild prediction scores up to 87.5\%.

\begin{figure*}[t]
    \parbox{\linewidth}{\centering\includegraphics[width=\linewidth]{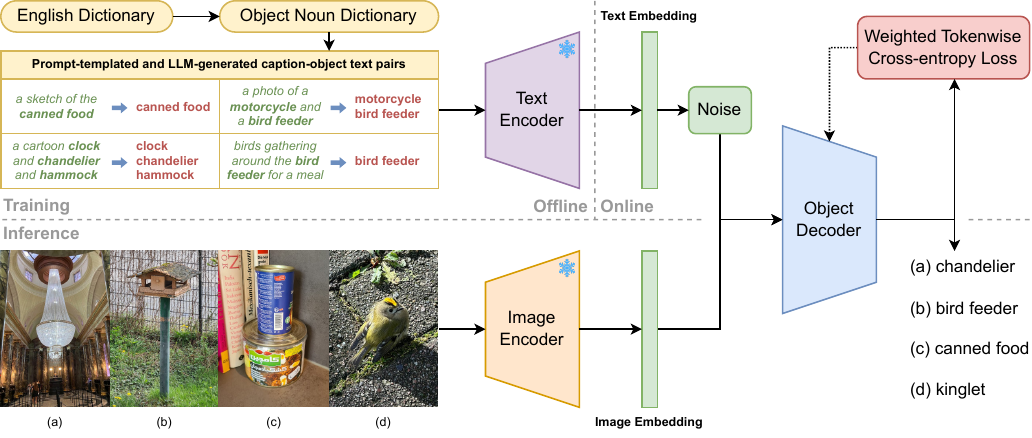}\vspace*{-6pt}}
    \caption{\textbf{Overview of the NOVIC training and inference schemes.} A dataset of caption-object text pairs is generated offline using an English dictionary, prompt templates, and LLM-based caption generation. The captions are encoded offline into text embeddings, and augmented with noise to train the object decoder. The training scheme seamlessly generalizes across the large modality gap between image and text embeddings, allowing inference of arbitrary images via the image encoder. Produced classifications can be very fine-grained.}
    \label{fig:architecture}
\end{figure*}

\section{Related Work}
\label{sec:related_work}

Learning a joint embedding space for multiple modalities such as images and text via contrastive pretraining produces powerful representations without the need for expensive annotation of datasets \cite{li_multimodal_2023,girdhar_imagebind_2023}. VLMs such as CLIP~\cite{radford_learning_2021} and ALIGN~\cite{jia_scaling_2021} paved the way for learning a broad range of visual concepts that can be transferred to downstream discriminative tasks, such as classification, via prompting \cite{gu_systematic_2023}. Given an image, a set of candidate categories, and prompt templates like \emph{``a photo of a [...]''}, CLIP can zero-shot classify the image by maximizing the similarity score between its encoded representation and all encoded candidate prompts. NOVIC, by contrast, frames the image classification problem as an unconstrained generative one, eliminating the need for category candidates and inference-time prompting. Various approaches build upon and extend the idea behind CLIP towards better pretraining \cite{sun_eva-clip_2023,zhai_sigmoid_2023,li_clipa-v2_2023} or fine-tuning \cite{wortsman_robust_2022} techniques, or add additional modalities such as audio \cite{guzhov_audioclip_2022,wu_wav2clip_2022}, video \cite{xu_videoclip_2021,yuan_florence_2021,huang_clover_2023,luo_clip4clip_2022,alayrac_flamingo_2022}, or point clouds \cite{xue_ulip_2023,zhang_pointclip_2022}. Other approaches use VLMs like CLIP as feature extractors for subsequent fine-tuning on more challenging downstream tasks such as visual question-answering \cite{shen_how_2022,song_clip_2022,eslami_pubmedclip_2023}, captioning \cite{mokady_clipcap_2021,barraco_unreasonable_2022}, motion planning \cite{shridhar_cliport_2022,tevet_motionclip_2022}, and navigation \cite{khandelwal_simple_2022}.

\emph{Open vocabulary learning} is a paradigm that emerged in the context of fine-tuning pretrained VLMs for vision tasks. It aims to overcome the closed-set assumption of standard classification models by allowing them to handle novel object categories outside of their training domain. Models such as VILD \cite{gu_open-vocabulary_2022}, RegionCLIP \cite{zhong_regionclip_2022}, and various successor models \cite{wu_clim_2024,wu_cora_2023,cheng_yolo-world_2024,pham_lp-ovod_2024,kaul_label_2022,du_learning_2022} enable VLMs to solve tasks such as open vocabulary object detection \cite{zareian_open-vocabulary_2021}, where, given an arbitrary set of candidate object labels, the model has to detect the corresponding objects in the image. Other works extend the open vocabulary paradigm to closely related visual tasks such as semantic segmentation \cite{he_clip-s4_2023,xu_open-vocabulary_2023,liang_open-vocabulary_2023,liu_open-world_2022,kirillov_segment_2023,lai_lisa_2024} and image tagging \cite{li_blip_2022,zhang_recognize_2024,huang_tag2text_2024}. Despite their merit, all the approaches mentioned above rely on candidate labels for inference, a strong assumption that we abandon in this work.

A parallel line of research aims to leverage the powerful representations provided by VLMs for generative tasks such as image captioning. As VLMs like CLIP cannot inherently produce sequences, approaches in this research area rely on additional decoders. Several methods have been proposed that either establish a mapping between a VLM and a pretrained text decoder \cite{mokady_clipcap_2021,fei_transferable_2023,tewel_zerocap_2022}, or train a decoder from scratch \cite{barraco_unreasonable_2022,kang_noise-aware_2023,shen_how_2022,cornia_generating_2023} using curated captioning datasets such as MS COCO~\cite{chen_microsoft_2015}, CC3M~\cite{sharma_conceptual_2018}, and Visual Genome~\cite{krishna_visual_2017}.

A recent line of research on zero-shot image captioning or retrieval with VLMs proposes text-only training to eliminate the need for curating expensive image datasets. As text and images are encoded into disjoint regions of the VLM embedding space, additional mechanisms to bridge the modality gap have been proposed, like injecting noise into the text embeddings \cite{nukrai_text-only_2022,gu_close_2023,gu_language-only_2024} or projecting the image embeddings \cite{li_decap_2023}. These methods face two significant disadvantages---first, captions often fail to capture key image aspects or identify core objects, and second, they are biased towards the limited set of object classes contained in their training dataset. NOVIC adopts text-only training, but focuses on predicting truly open vocabulary object labels.

\section{Methodology}
\label{sec:methodology}

An overview of the proposed training and inference schemes for NOVIC is shown in \Cref{fig:architecture}. A text-only dataset mapping caption sentences to target object nouns is synthetically generated offline (cf.~\Cref{sec:training_dataset}). The text embedding vectors of the captions are then precomputed using the text encoder of a frozen CLIP model. These vectors are augmented online with unit-norm-preserving noise, and used to train a decoder-only transformer, the \emph{object decoder} (cf.~\Cref{sec:decoder_model}), to generate the object noun corresponding to each caption. The output object nouns are generated in the form of free language using the same text tokenizer as the frozen text encoder, with no definition of candidate object nouns being provided to the decoder in any way. At inference time, the object decoder seamlessly generalizes to image embedding vectors calculated using the image encoder of the frozen CLIP model, despite CLIP models generally having very large modality gaps \cite{liang_mind_2022}. The performance of the decoder is measured on image classification datasets and annotated open vocabulary image datasets (cf.~\Cref{sec:evaluation_datasets}).

\subsection{Object Decoder Model}
\label{sec:decoder_model}

\begin{figure*}[t]
    \parbox{\linewidth}{\centering\includegraphics[width=0.9\linewidth]{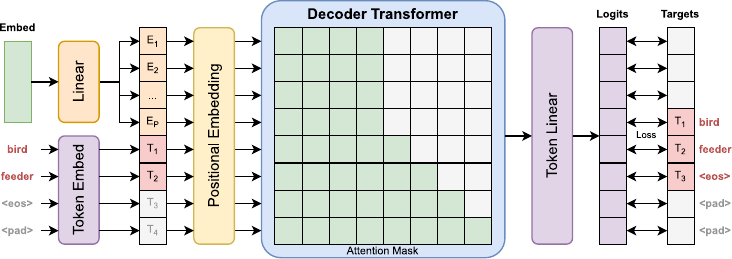}}
    \caption{\textbf{Architecture of the object decoder.} The decoder-only transformer is given a sequence of linearly projected embedding vector tokens (full attention) followed by autoregressive object noun tokens (causal attention). The output sequence is converted to token logits using a linear layer that is weight-tied to the token embeddings. Cross-entropy loss with left-shifted target tokens is used during training.}
    \label{fig:object_decoder}
    \vspace*{-1ex}
\end{figure*}

The architecture of the object decoder model is shown in \Cref{fig:object_decoder}. The model is built around an autoregressive decoder-only language transformer \cite{vaswani_transformers_2017} that has been configured to use learned token embeddings, weight tying \cite{press_weighttying_2016}, learned positional embeddings, pre-layer normalization \cite{xiong_preln_2020}, GELU activation functions \cite{hendrycks_gelu_2016}, beam search, custom weight initialization (cf.~\Cref{app:weight_init}), a contractive inner feed-forward dimension, and no biases in the linear and layer normalization layers, including the final token logits linear layer. Dropout is applied after the positional embedding and within the transformer layers. The input embedding vector is supplied to the transformer in the form of $P$ prefix token vectors, \ie $P$ dynamically calculated vectors of dimension $H$ (hidden dimension). The prefix tokens are computed from the input embedding using a learned linear projection that has $PH$ outputs and no biases, as the immediately following positional embedding makes biases redundant. The tokenized target object nouns do not require a start token, as the first output token is predicted at the sequence location of the last prefix token. Appropriate sequence masking is also used to ensure that only sequence locations up to the first output \texttt{<eos>} contribute to the loss. The internal attention mask of the transformer is causal, with the exception that the $P$ prefix tokens can attend to each other without restriction, allowing them to share information freely.

\subsection{Training Dataset for Text-Only Supervision}
\label{sec:training_dataset}

The object decoder needs to be trained on a dataset of associative caption-object text pairs (cf.~\Cref{fig:architecture}). One could attempt to generate such a dataset from a large-scale image-text web dataset ranging into the billions of samples, but it is not a simple task to accurately and sufficiently efficiently construct the required ground truth object noun annotations, especially in such a way that a fair coverage of the English language is achieved. We instead \emph{invert} the chosen CLIP text encoder by synthetically generating captions that explicitly cover the entire English language of object nouns, and use this to systematically sample in the CLIP embedding space essentially all possible object concepts, and learn to map them to their corresponding textual representations.

\paragraph{Object noun dictionary.}
An exhaustive dictionary of simple and compound object nouns, referred to as the \emph{object noun dictionary}, was created based on the WordNet dictionary \cite{miller_wordnet_1995}, GNU Collaborative International Dictionary of English \cite{dict_gcide}, and the categories used by various image datasets (cf.~\Cref{app:object_noun_dict}). Noun entries were cleaned, collected and merged as appropriate to account for word duplication, plurality, inflection and alternate spellings, and filtered down to only include nouns that correspond to suitable object classifications (\eg no abstract nouns). This resulted in \numprint{42919} unique case-normalized object nouns corresponding to \numprint{96526} noun variants. Each object noun was then assigned a log-scale frequency value based on its observed frequency in the English language as per the n-gram frequencies in the Google Web Trillion Word Corpus \cite{dict_ngrams,dict_trillion}. Object nouns that are categories in a curated, comprehensive collection of image datasets were given appropriate frequency boosts. The log-scale frequencies were scaled and rounded so that very rare nouns saturate at a value of 1, and very frequent nouns have values $\geq$12. Aside from being used to adjust relative sampling frequencies during training, the resulting frequencies allowed the construction of object noun dictionary subsets by filtering out nouns less or equal to a particular frequency threshold (FT), \eg FT2.\footnote{FT0, FT2, FT6, and FT9 have \numprint{42919}, \numprint{11897}, \numprint{5899}, and \numprint{2919} unique object nouns, respectively. FT0 is the complete object noun dictionary.}

\paragraph{Prompt templates.}
A direct and extremely cheap way of generating caption-object text pairs is by using exactly the same prompt templates as is usually used for zero-shot CLIP classification. This maps captions like \emph{``a blurry photo of a bottle''} to the object noun \emph{``bottle''}. Adjusting the ImageNet-1K and CIFAR prompts used by \cite{radford_learning_2021} to account for pluralization and allow for correct indefinite article use leads to 85 unique prompt templates for singular forms of object nouns, and 68 for plural forms. Combined with the full object noun dictionary, and including repetitions based on the log-scale frequencies of the nouns leads to 13.4M samples in the dataset. The mean observed object noun frequency is 3.5, but the true mean number of repetitions is much smaller at only 1.5, due to the use of the available alternate spellings during prompt templating.

\vspace{-1ex}
\paragraph{Multiset prediction.}
Image embeddings can capture significantly more complicated dynamics than just the main entity present. If there is a photo of a lion with a zookeeper, for instance, the image embedding will likely contain some features corresponding to \emph{lion}, \emph{person}, \emph{fence}, and potentially more abstract concepts like \emph{zoo} as well. To help also capture these dynamics in the text embeddings seen during training, multi-target samples are introduced. The complete object noun dictionary, including explicit repetition of alternate spellings based on their individual log-scale frequencies of mean 1.5, is randomly shuffled $m$ times into $m$ separate lists. These lists are `zipped' to form $m$-tuples of object nouns that are then used to generate prompt-templated captions exactly as before. The $m$ object nouns per caption are linked by `and' (cf.~\Cref{fig:architecture}), and all $m$ nouns use a target weight of $\tfrac{1}{m}$. This produces a dataset of 13.4M further samples for each value of $m$, referred to as the $\mathcal{M}_m$ \emph{multiset} in each case (\eg $\mathcal{M}_3$ for $m = 3$). In total, a combined dataset of $\mathcal{M}_1$\footnote{$\mathcal{M}_1$ is equivalent to the prompt-templated dataset as initially described.} and the two multisets $\mathcal{M}_2$ and $\mathcal{M}_3$ contains 40.2M samples with 80.4M targets. We qualitatively observed that multiset training specifically helped the open vocabulary classification of images that prominently contain multiple different elements or objects. On an intuitive level, the use of multi-target samples also prevents the object decoder from being able to just `rote-memorize' which local regions of the text embedding space correspond to which pure object concepts, and instead forces the decoder to learn \emph{which} values in \emph{which} parts of the embedding vector are actually associated with \emph{which} object concept. Such knowledge is essential for proper generalization to image embeddings.

\paragraph{Caption data.}
In order to supply the object decoder with even more diverse synthetic training data to generalize from, captions were generated for each noun in the object noun dictionary using an LLM (cf.~\Cref{app:caption_gen}). This approach leverages the implicit contextual knowledge of LLMs to place object nouns in \emph{relevant} contexts---incorporating appropriate adjectives, actions, scenes, and secondary objects---in a style resembling real image captions.\footnote{CLIP embedding spaces are \emph{trained} on real image captions.} The minimum number of unique generated captions for each noun variant was set proportionally to the corresponding log-scale frequency (up to 100) in order to match the relative word frequencies found in the English language. OpenAI GPT-3.5 Turbo was selected as an ideal balance of cost and sufficient performance, and was used to generate a total of 1.8M captions. Ensuring frequency-appropriate use of alternate spellings, this resulted in a total of 2.9M caption samples (2.0M unique) that were then oversampled to around 30\% the size of any multiset they were merged with. For the combined dataset $\mathcal{M}_{1-3}$ this resulted in 51.7M samples with 91.9M targets.

\paragraph{Noise augmentation.}
In order to allow more robust and generalizable features of the CLIP embedding space to be learned, data augmentation in the form of noise is used during training. Due to the large modality gap between text and image embeddings \cite{liang_mind_2022}, noise is in fact a crucial component of the training scheme. Noise allows the object decoder to learn to be invariant to even very large changes in the non-object-related parts of the text embedding vectors, which helps the generalization to image embeddings, as these are very sensitive to secondary objects, backgrounds, and even subtle visual variations. Gu et al\@. \cite{gu_close_2023} used elementwise Gaussian noise with vector renormalization to train a captioning model on text embeddings only, but did not entirely accurately explain its effect. The CLIP embedding space is deceptively and unimaginably vast\footnote{Even if we assumed each element of a 768-dim embedding vector only contains a \emph{single} bit of information, this would still lead to $10^{231}$ different embeddings, against which even a trillion samples is essentially zero.}, so the random noise statistically \emph{cannot} be significantly filling the gaps between samples, or causing a meaningful overlap with image embedding vectors. The main benefit of noise is that the model can learn to focus on which parts (\ie subspaces) of the embedding vector really encode which individual object concepts, instead of focusing on which region of the embedding space a vector is in as a whole. We observe that the amount of noise required for best training results is actually quite large, not small or minor \cite{gu_close_2023}, and its effect is to completely scatter the cone \cite{liang_mind_2022} of text embeddings to a thin hyperannulus on the order of 66--80\degree away, with no single sample remaining significantly closer than that to any original text embedding (cf.~\Cref{app:noise_effect}). By comparison, the very large modality gap between the image and text embedding cones can also be measured, and is around 70--85\degree for \emph{matching} image-text pairs (depending on the CLIP model), and only 8--12\degree more than this for completely uncorrelated image-text pairs. As indicated previously though, statistically there is negligible region overlap between the scattered hyperannulus and the image embeddings cone due to the vast nature of the embedding space. The large amount of noise required can instead be seen as a trade-off between the disruptive `blurring' effect of noise, and the benefit of the object decoder having seen many embedding vectors during training that are as far removed from text embeddings as image embeddings are. In this work, we use an 85\% to 15\% mix of elementwise Gaussian noise and uniform angle noise. The latter was developed to beneficially cover a greater range of noise dynamics and levels, by rotating input text embedding vectors uniformly in random directions by angles in the uniform 45--75\degree range.

\subsection{Evaluation Datasets for Zero-Shot Recognition}
\label{sec:evaluation_datasets}

One way to evaluate how much NOVIC has learned is by examining its zero-shot performance on standard image classification datasets like ImageNet-1K and Food-101. Context is everything for image classification datasets however. Knowing which categories are available in a dataset serves as a \emph{strong} prior for supervised models. For instance, if a dataset contains only one bird category, such as \emph{kinglet}, a supervised model only needs to determine whether an image is a bird to confidently classify it as a kinglet. For prompt-free open vocabulary models like NOVIC however, the task is more challenging. When presented with an image that happens to come from Food-101, it is not clear that the model should suddenly focus on the specific dish shown instead of potentially more prominent or general objects in the image. Consequently, to meaningfully evaluate NOVIC on image classification datasets, we restrict the generated beam search text tokens in every forward pass of the object decoder to those consistent with the defined category names, and score whether the correct final category is identified.

As NOVIC is trained without providing explicit categories, we wish to evaluate it also on datasets that do not specify categories. No suitable such dataset is known however, as such a dataset would need to have annotations for every possible correct object noun in the English language for every visible entity in every part of each image. To address this, three open vocabulary image datasets were curated as part of this work (cf.~\Cref{app:open_vocab_datasets}), and individually annotated by both human and multimodal LLM annotators (cf.~\Cref{app:llm_annotation}) for the object nouns that were predicted by the trained models. The annotations specify whether each classification is \emph{correct}, \emph{close}, or \emph{incorrect}, and for the human annotations, whether it relates to a \emph{primary} or \emph{secondary} element of the image. The three new datasets are (i) World: 272 images of which the grand majority are originally sourced (have never been on the internet) from 10 countries by 12 people, with an active focus on covering as wide and varied concepts as possible, including unusual, deceptive and/or indirect representations of objects, (ii) Wiki: \numprint{1000} Wikipedia lead images sampled from a scraped pool of 18K, (iii) Val3K: \numprint{3000} images from the ImageNet-1K validation set, sampled uniformly across the classes (LLM annotations only). The suffixes -H and -L are used to clarify whether human or LLM annotations are in use, \eg Wiki-H. A total of 17.4K human and 112K LLM (OpenAI GPT-4o) annotations were made across the three datasets. This was at the limit of feasibility in terms of number of samples, person-hours and cost.


\section{Experiments}
\label{sec:experiments}

For experiments, the SigLIP ViT B/16 model \cite{zhai_sigmoid_2023} is used as the base CLIP model, due to its favorable balance of speed and performance, especially when compared to the original CLIP models \cite{radford_learning_2021}. Training is performed for 18 epochs with the AdamW optimizer, cosine learning rate decay, weight decay on all multidimensional weights, and a gradient-accumulated batch size of $B = 8192$. Hyperparameter details can be found in \Cref{app:hyperparameters}. The object decoder model is quite small at only 12.2M parameters, and takes less than 3 days to train on a \emph{single} RTX A6000. For FT2, this drops to 11.4M parameters and only 1.5 days. Overall, this is \emph{extremely} efficient, considering that up to a billion samples are being seen by a single GPU. As a scale comparison, using our method on the very same GPU to train an object decoder on all 1.28M ImageNet-1K samples for 18 epochs takes a mere 19 minutes, and results in an impressive top-1 accuracy of 88.21\%. The object decoder is also very resource efficient at inference time. Requiring at most 5\,GB GPU memory, the total pipeline---including CLIP model, autoregression, and beam search---takes on average 26\,ms per image for single images, and 7\,ms per image when using batching ($B = 256$), making it very real-time capable.

\paragraph{Open vocabulary scoring.}
Model predictions on the three introduced open vocabulary image datasets are scored by assigning \{1, 0.8, 0.5, 0.4, 0\} points for \{\emph{correct primary}, \emph{correct secondary}, \emph{close primary}, \emph{close secondary}, \emph{incorrect}\} predictions respectively, and expressing the sum as a percent of the maximum possible score. For verification purposes, the prediction scores achieved by many models were statistically compared between human and LLM annotations on the same dataset and found to be suitably correlated (cf.~\Cref{app:llm_annotation}), thus allowing model performance insights to be gained even when manual annotation is just too prohibitive. It is expected that LLM prediction scores are lower than human ones due to some residual LLM mistakes, a generally narrow acceptance of classifications, and inherently less frequent use of the \emph{close} annotation by the LLM.

\subsection{Zero-Shot Classification Performance}
\label{sec:zero_shot_perf}

\newlength{\daggerwidth}
\begin{table}[t]
    \centering\small
    \setlength\tabcolsep{1.3pt}
    \settowidth{\daggerwidth}{${}^\dagger$}
    \begin{tabular}{l>{\centering\arraybackslash}p{1.0cm}|>{\centering\arraybackslash}p{1.0cm}>{\centering\arraybackslash}p{1.0cm}>{\centering\arraybackslash}p{1.0cm}>{\centering\arraybackslash}p{1.0cm}}
        \toprule
        Dataset & CLIP & FT9 & FT6 & FT2 & FT0 \\
        \midrule
        ImageNet-1K \cite{imagenet1k} & \emph{75.91} & 46.76${}^\dagger$\hspace*{-\daggerwidth} & \textbf{69.50} & 68.11 & 60.56 \\
        Tiny ImageNet \cite{tinyimagenet} & \emph{59.77} & 54.53${}^\dagger$\hspace*{-\daggerwidth} & \textbf{57.36} & 55.83 & 53.78 \\
        Imagenette \cite{imagenette_woof} & \emph{99.59} & \textbf{99.44} & 99.27 & 99.32 & 99.23 \\
        Imagewoof \cite{imagenette_woof} & \emph{93.28} & \textbf{93.07} & 92.07 & 92.20 & 90.15 \\
        ImageNet-A \cite{hendrycks_nae_2021} & \emph{45.05} & 35.33${}^\dagger$\hspace*{-\daggerwidth} & \textbf{42.43} & 41.42 & 38.23 \\
        ImageNet-R \cite{hendrycks_nae_2021} & \emph{90.24} & 74.46${}^\dagger$\hspace*{-\daggerwidth} & \textbf{88.01} & 86.62 & 83.43 \\
        Food-101 \cite{bossard_food-101_2014} & \emph{91.55} & \textbf{85.25} & 83.24 & 82.73 & 68.80 \\
        CIFAR-10 \cite{krizhevsky_learning_2009} & \emph{92.33} & \textbf{91.08} & 90.53 & 90.77 & 90.92 \\
        CIFAR-100 \cite{krizhevsky_learning_2009} & \emph{72.19} & \textbf{70.65} & 69.93 & 69.43 & 67.80 \\
        \bottomrule
    \end{tabular}
    \caption{\textbf{Performance on image classification benchmarks (\% mean of 3).} ${}^\dagger$FT9 has disadvantaged scores on some datasets due to the seen object nouns not covering all the classes (cf.~\Cref{app:cls_perf_discussion}).}
    \label{tab:image_cls}
\end{table}

\begin{table}[t]
    \centering\small
    \setlength\tabcolsep{1.3pt}
    \begin{tabular}{p{2.95cm}>{\centering\arraybackslash}p{1.0cm}>{\centering\arraybackslash}p{1.0cm}>{\centering\arraybackslash}p{1.0cm}>{\centering\arraybackslash}p{1.0cm}}
        \toprule
        Annotations & FT9 & FT6 & FT2 & FT0 \\
        \midrule
        ImageNet-1K & 46.52 & 69.47 & 67.92 & 60.72 \\
        Open (correct) & \textbf{70.33} & \textbf{75.04} & \textbf{72.90} & \textbf{66.19} \\
        \midrule
        Open (correct\,+\,close) & \emph{71.90} & \emph{76.50} & \emph{74.35} & \emph{67.86} \\
        \bottomrule
    \end{tabular}
    \caption{\textbf{Performance on Val3K dataset (\% mean of 3),} showing that category-constrained scores are a significant underestimation of the true open vocabulary performance of NOVIC.}
    \label{tab:image_cls_val3k}
\end{table}

\paragraph{Image classification benchmarks.}
\Cref{tab:image_cls} shows the zero-shot performance of NOVIC on various image classification benchmarks. The performance of the underlying CLIP model is also shown as a measure of how completely the vast knowledge of the CLIP model has been captured by the object decoder and converted into generative textual understanding. The decoder relies on the organization of the embedding space provided by the CLIP model, and so naturally is not expected to outperform it in this category-prompted setting. Overall, NOVIC shows competitive top-1 accuracies, with some `distillation' losses being observed for tendentially finer-grained classification datasets and lower frequency thresholds. Classification benchmarks, however, do not tell the whole story. The act of restricting the object decoder to a fixed vocabulary \emph{forces} it to choose object nouns it does not even consider to be best. In \Cref{tab:image_cls_val3k}, we evaluate the top-1 accuracy on Val3K using the original ImageNet-1K annotations, and compare this to the percent of open vocabulary classifications on the very same images that are \emph{correct} as per Val3K-L. Prediction scores that include points for \emph{close} predictions are also shown. The results demonstrate that the object decoder's knowledge is more accurately assessed when used for unconstrained open vocabulary classification.

\begin{figure}[t]
    \parbox{\linewidth}{\centering\includegraphics[width=\linewidth]{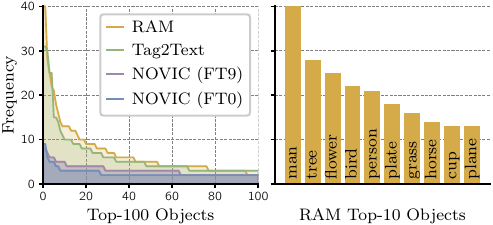}\vspace*{-3pt}}
    \caption{\textbf{Diversity comparison of predicted object nouns.} Plots showing the sorted top frequency counts of the object nouns predicted for Wiki-H. Even when trained on FT9, NOVIC shows much greater diversity and has a less peaked frequency distribution. \textbf{Right:} The top-10 nouns for RAM are generic and overused.}
    \label{fig:frequencies}
    \vspace*{-1.0ex}
\end{figure}

\paragraph{Comparison with related work.}
As discussed in \Cref{sec:related_work}, no directly equivalent state-of-the-art model is known that performs real-time prompt-free open vocabulary image classification. The most suitable models for comparison are image tagging models like Tag2Text \cite{huang_tag2text_2024} and the Recognize Anything Model (RAM) \cite{zhang_recognize_2024}, as they are designed to be able to provide text labels for images of arbitrary content. We evaluate the largest model variants by taking the highest scoring prediction for each image, and compare this to NOVIC in \Cref{tab:tagging_model_comparison,app:additional_results}. While the tagging models at first glance numerically seem to obtain a reasonable \emph{Prediction Score}, the comparatively low \emph{Primary Scores} indicate that the models, unlike NOVIC, make a large portion of their score from \emph{secondary} or \emph{close} predictions, as opposed to \emph{correct primary} ones (a deficiency clearly illustrated in \Cref{tab:tagging_model_comparison_world}). Also, when accounting for prediction specificity, the resulting \emph{Overall Score} clearly demonstrates the superiority of NOVIC by up to 19\% absolute. A closer look reveals that the low scores of the tagging models are due to a severe lack of diversity and specificity (cf.~\Cref{fig:frequencies} and qualitative results in \Cref{app:qualitative_comparison}). The produced tags are tendentially coarse and repetitive, with around a third of all images receiving one of only 20 generic top labels, often missing the main point of images. Even when provided with the full object noun dictionary, the total diversity of labels stays below 400, with over 80\% of predictions still being made from the small core RAM vocabulary. This is in contrast to our model, which has a much flatter prediction distribution, diversities up to 934 (cf.~\Cref{tab:tagging_model_comparison}), and a remarkably high prediction score despite generating \emph{fine-grained} classifications. The tags produced by Tag2Text and RAM also tend to be uninformative---when applied to the images of the Imagewoof \cite{imagenette_woof} validation set for example, both models simply output \emph{dog} 88\% of the time, and never a single dog breed. Our FT0 model, on the other hand, predicts over 90 different dog breeds, with much of the weight covering variations of the actual ground truth classes. When RAM is evaluated on ImageNet-1K by only providing it the 1000 class names to choose from, it only scores 38.3\% (25.4\% for non-core vocabulary), which is \emph{far} less than the 75.54\% of the ViT-L CLIP model it was trained on, and \emph{far} less than the 60--70\% nominally scored by NOVIC.

\begin{table}[t]
    \centering\footnotesize
    \setlength\tabcolsep{1.3pt}
    \begin{tabular}{p{1.85cm}>{\centering\arraybackslash}p{0.95cm}>{\centering\arraybackslash}p{0.95cm}>{\centering\arraybackslash}p{0.75cm}>{\centering\arraybackslash}p{1.0cm}>{\centering\arraybackslash}p{1.2cm}>{\centering\arraybackslash}p{0.95cm}}
        \toprule
        Model & Seen Objects & Used Objects & RAM Vocab & Primary Score & Prediction Score & Overall Score \\
        \midrule
        Tag2Text \cite{huang_tag2text_2024} & \numprint{3429} & 397 & 89.0 & 69.15 & \textbf{81.35} & 62.61 \\
        RAM \cite{zhang_recognize_2024} & \numprint{4585} & 348 & 100.0 & 65.05 & 80.73 & 59.98 \\
        RAM (FT2) & \numprint{11897} & 391 & 83.4 & 62.30 & 77.62 & 60.58 \\
        RAM (FT0) & \textbf{\numprint{42919}} & 399 & 81.7 & 61.70 & 76.94 & 60.25 \\
        NOVIC (FT9) & \numprint{2919} & 693 & 64.0 & 68.55 & 72.35 & 69.08 \\
        NOVIC (FT6) & \numprint{5899} & 794 & 49.3 & 69.90 & 72.70 & 71.07 \\
        NOVIC (FT2) & \numprint{11897} & 846 & 39.7 & 68.65 & 72.53 & 71.88 \\
        NOVIC (FT0) & \textbf{\numprint{42919}} & 852 & 37.5 & 68.70 & 72.34 & 71.31 \\
        NOVIC${}^\dagger$ (FT2) & \numprint{11897} & 893 & 36.3 & 77.05 & 80.13 & \textbf{79.02} \\
        NOVIC${}^\dagger$ (FT0) & \textbf{\numprint{42919}} & \textbf{934} & \textbf{28.7} & \textbf{77.10} & 79.18 & 78.21 \\
        \bottomrule
    \end{tabular}
    \caption{\textbf{Open vocabulary Wiki-H results in comparison to baselines (best of 3).} Tag2Text and RAM achieve artificially high scores by frequently outputting uninformative generic tags like \emph{man}, \emph{tree}, \emph{grass}, \emph{plate}, and not for instance what is \emph{on} the grass or plate, or what \emph{type} of tree it is, or what the man is \emph{using}. Scores are in \%. \emph{RAM Vocab} is the percent of samples whose predictions are in the core RAM vocabulary. \emph{Primary Score} is the \emph{Prediction Score} when rewarding only primary predictions. \emph{Overall Score} is the \emph{Prediction Score} with $\times$0.5 applied to the scores of coarse predictions. See \Cref{app:additional_results}. ${}^\dagger$Using DFN-5B H/14-378 CLIP model.}
    \label{tab:tagging_model_comparison}
\end{table}

\subsection{Ablation Studies}
\label{sec:ablations}

\Cref{tab:ablations} shows the effect on decoder performance of varying the dataset configuration, training noise, and underlying CLIP model. FT2 was chosen as the ablation baseline, as many ablation runs were required, and FT2 strikes a balance between overall feasibility of training times, and ensuring that a diverse and expressive range of object nouns are seen during training. Further ablations are presented in \Cref{app:decoder_arch_ablations}. Overall, as expected, the human annotations exhibit generally higher scores and are slightly more consistent, due to human checking and more prevalent use of the \emph{close} category. The decoder ImageNet-1K scores are also observed to be well-correlated with general open prediction performance.

\paragraph{Training dataset.}
The effect of the object noun dictionary frequency threshold (cf.~\Cref{sec:training_dataset}) on decoder classification performance is shown in \Cref{tab:ablations}. The prediction scores grow as the training dataset size is increased, until a limit is reached between FT2 and FT0 where the exploding diversity of seen object nouns saturates the amount of order the CLIP model can provide in the embedding space. In \Cref{tab:ablations}, we systematically analyze the contribution of each component of the training set. Starting with $\mathcal{M}_1$ and incrementally building the nominal training set (\ie $\mathcal{M}_1$ + $\mathcal{M}_2$ + $\mathcal{M}_3$ + LLM captions), we observe consistent gains. Immediate losses are seen if the LLM captions are removed, or if the lower-order $\mathcal{M}_1$ and $\mathcal{M}_2$ multisets are removed. Further tests with variations in training noise configuration also validate that very large amounts of Gaussian noise are \emph{required} in order to produce models that can generalize across the modality gap,\footnote{\emph{Gauss 3.25} in the table refers to elementwise Gaussian noise with a standard deviation so high that the expected norm of the noise vector that is added to the unit-norm embedding is 3.25, \ie many times its magnitude.} and that instead using uniform angle noise on 15\% of the samples produces the best seen configuration.

\begin{table}[t]
    \centering\small
    \setlength\tabcolsep{1.3pt}
    \begin{tabular}{l@{\hspace{0.8em}}>{\centering\arraybackslash}p{1.12cm}>{\centering\arraybackslash}p{1.12cm}>{\centering\arraybackslash}p{1.12cm}>{\centering\arraybackslash}p{1.12cm}}
        \toprule
        Configuration & World-H & Wiki-H & Wiki-L & IN1K \\
        \midrule
        FT9 & 76.13 & 71.20 & 63.57 & 46.76 \\
        FT6 & 78.42 & \textbf{72.33} & 62.62 & \textbf{69.50} \\
        FT2* & \textbf{78.92} & 72.03 & \textbf{63.97} & 68.11 \\
        FT0 & 73.98 & 71.48 & 61.27 & 60.56 \\
        \midrule
        $\mathcal{M}_1$ dataset & 70.11 & - & 50.05 & 63.88 \\
        \;\;+ LLM captions & 72.16 & - & 55.37 & 66.06 \\
        \;\;+ $\mathcal{M}_2$ dataset & 78.37 & - & 61.60 & 67.97 \\
        \;\;+ $\mathcal{M}_3$ dataset* & \textbf{78.92} & \textbf{72.03} & \textbf{63.97} & \textbf{68.11} \\
        $\mathcal{M}_1$ + $\mathcal{M}_2$ + $\mathcal{M}_3$ & 77.56 & 71.60 & 62.08 & 65.36 \\
        $\mathcal{M}_3$ + LLM captions & 77.88 & 71.28 & 62.93 & 61.12 \\
        \midrule
        Gauss 1.00 & 19.44 & - & 15.62 & 13.36 \\
        Gauss 1.75 & 70.37 & - & 52.07 & 49.23 \\
        Gauss 2.50 & 76.26 & - & 62.10 & 67.94 \\
        Gauss 3.25 & 77.57 & 70.95 & 62.38 & 66.46 \\
        \;\;+ 15\% Uniform 45--75* & \textbf{78.92} & \textbf{72.03} & \textbf{63.97} & \textbf{68.11} \\
        Uniform 45--75 & 76.46 & - & 63.27 & 64.75 \\
        \midrule
        DataComp-1B B/16 \cite{gadre_datacomp_2023} & 76.85 & - & 63.55 & 64.17 \\
        SigLIP B/16* \cite{zhai_sigmoid_2023} & 78.92 & 72.03 & 63.97 & 68.11 \\
        DataComp-1B L/14 \cite{gadre_datacomp_2023} & 84.27 & - & 70.35 & 70.05 \\
        SigLIP SO/14 \cite{zhai_sigmoid_2023} & \textbf{87.49} & - & 70.28 & 74.85 \\
        DFN-5B H/14-378 \cite{fang_dfn_2024} & 87.08 & - & \textbf{74.88} & \textbf{75.82} \\
        \;\;$\rightarrow$ FT0 & 86.41 & - & 74.48 & 75.68 \\
        \bottomrule
    \end{tabular}
    \caption{\textbf{Dataset, training noise, and CLIP model ablations (\% mean of 3).} IN1K is the ImageNet-1K classification performance of the trained decoders, and the asterisks correspond to the nominal ablation baseline. See \Cref{app:decoder_arch_ablations} for more ablation results. Some Wiki-H scores are unavailable due to the high cost of manual annotation.}
    \label{tab:ablations}
    \vspace*{-2ex}
\end{table}

\vspace*{-1.5ex}
\paragraph{CLIP model.}
The proposed NOVIC training scheme scales well with the performance of the underlying CLIP model. \Cref{tab:ablations} demonstrates that progressively stronger CLIP models yield progressively stronger object decoders. Remarkably however, the training times remain almost unchanged, as text encoders only scale slowly with CLIP model strength and the datasets of text embedding vectors are generated offline and only once anyway (cf.~\Cref{fig:architecture}). It can be observed from \Cref{tab:ablations} that the performance of the FT0 object decoder that was trained on the strong DFN-5B CLIP model is only marginally below the corresponding FT2 performance. This is much closer than for the nominal CLIP model, and can be attributed to the improved order and separation of concepts that the CLIP model provides in the embedding space. Based on the World-H results, it is actually expected that the DFN-5B FT0 model can provide very fine-grained \emph{correct primary} classifications for up to 84.2\% of pictures it is given. Having trained on the full unfiltered object noun dictionary, the model is frequently able to make correct classifications as detailed as \emph{red velvet cake}, \emph{brazil nut tree}, \emph{electron microscope}, and \emph{egretta caerulea}.

\vspace*{-1.5ex}
\paragraph{Language generalization.}
During training, the object decoder develops an implicit understanding of language and its relationship to object concepts, demonstrating qualitative evidence that it can occasionally use this knowledge to generalize beyond the dataset. For instance, the unseen noun \emph{New Zealand willow} was predicted for a scenic photo of a willow tree that was indeed taken in New Zealand. The object decoder also inherits CLIP's slight tendency to be able to `read', as evidenced for example by a generic blue plushie being correctly classified by multiple decoders as \emph{genus staphylococcus}, but not anymore if the small visible text \emph{Staphylokokke} is removed, falling back to \emph{stuffed toy}. In another case, a sign reading \emph{Aral} is classified as \emph{ara} (a macaw), indicating that the decoder can associate certain features in the embedding space with certain spellings. As another form of language generalization, the object decoder can easily be trained to output high-performing predictions in any chosen language---even low resource languages for which insufficient image-text data exists to train a CLIP model. By simply applying dictionary or machine translation to the target object nouns in the $\mathcal{M}_m$ multisets, a predominantly English CLIP model can be used to directly train a strong object decoder in any translatable language.

\section{Conclusion}
\label{sec:conclusion}

In this paper, we address the problem of unconstrained real-time open vocabulary image classification, and highlight the limitations of current state-of-the-art methods in accomplishing this task. To overcome these limitations, we propose an autoregressive object decoder trained on a large-scale text-only dataset. The decoder generates object nouns directly as language on the basis of CLIP embeddings, and crucially, does not require any class candidates to be provided. To create the text-only training dataset, we introduce a comprehensive object noun dictionary and employ multiset prompt-templating in combination with LLM-generated captions and a pivotal noise augmentation strategy to effectively bridge the modality gap from text to images. Our approach demonstrates promising results on both established and newly proposed in-the-wild classification datasets. We lift the common limitation of zero-shot image recognition models regarding prompts and class candidates, ultimately allowing agents to immediately recognize and label \emph{any} object in dynamic environments. Future work can include providing an input to the decoder that controls the level of semantic granularity used in the predicted output nouns.

{\small
\bibliographystyle{ieee_fullname}
\bibliography{egbib}}

\clearpage
\appendix

\section*{\Large Appendix}

\section{Object Decoder Weight Initialization}
\label{app:weight_init}

Various state-of-the-art transformer models like BERT, OpenCLIP and GPT use heuristics for weight initialization, for example by initialising weights to a zero-mean normal distribution of standard deviation 0.02 or $\tfrac{1}{\sqrt{H}}$, for $H$ the hidden dimension. Guided by the principles of mean field theory, we rigorously calculate weight initialization standard deviations such that the expected forward pass norms are statistically preserved throughout the object decoder, and such that the overall contributions of the main and residual pathways are balanced, avoiding excessive input signal dilution. This comes with numerical and gradient stability benefits, leading to fast and robust initial convergence.

\subsection{Strategies for Strong Initial Convergence}
\label{app:weight_init_strats}

In general, the rule-of-thumb aims of the presented \emph{variance-balanced weight initialization strategy} for the forward pass of a network are:
\begin{itemize}
    \item \textbf{Avoid input signal dilution.} For example, if the residual pathway of the input through the network involves repeatedly combining it with values calculated from random weights and renormalizing, then the `proportion' of the final output signal that actually corresponds statistically to the input signal can be exponentially microscopic, making initial convergence more difficult.
    \item \textbf{Avoid combining unbalanced scales of data.} If two signals are added or concatenated that have significantly different scales (multiplication is less critical), especially if done repeatedly, then this has potentially negative numerical effects on backpropagation. It can also quickly lead to signal dilution.
    \item \textbf{Avoid significant scale decay/growth.} For example, if each residual block in a network initially causes a rescaling of signals by only 50\% on average, then a moderate 12-layer transformer with cross-attention will initially produce outputs that are only one 70 billionth of the inputs, which is undesirable for learning.
    \item \textbf{Aim for around unit scale throughout.} Making an optimizer deal with vastly different scales of signals and gradients throughout the network is undesirable, even if the optimizer has features to help deal with it. Floating-point representations have their limits, especially if attempting to train in reduced or mixed precision. Certain layers, like for example certain activation functions (\eg GELU), are also `designed' for approximately unit standard deviation scales of input data.
\end{itemize}

Based on these rules of thumb, it can immediately be seen why transformers with post-layer normalization \cite{xiong_preln_2020} in general struggle to train for larger numbers of layers. In each residual block, the initially random weights in the residual path cause an essentially random residual signal to be added to the input signal, before the combined signal is then layer-normalized. This results in the input signal to the next residual block only being a certain fraction the input signal, and the rest just random noise. As this combined signal is now the new input signal for the next block, it further downscales the original input signal by a certain factor and adds even more random noise. This process continues throughout the entire transformer, leading to prototypical signal dilution, with the problem getting \emph{exponentially} worse for larger numbers of layers. This is why pre-layer normalization was chosen for NOVIC.

\subsection{Data Variance Propagation}
\label{app:weight_init_propagation}

We must first understand how variances propagate through the individual layers of a transformer network before we can integrate this knowledge into a complete strategy.

\paragraph{Addition.}
If a vector containing elements of variance $\sigma_1^2$ is added to an (uncorrelated) vector of variance $\sigma_2^2$, then the variance of the elements in the sum of the two vectors is
\begin{equation}
\sigma_{1+2}^2 = \sigma_1^2 + \sigma_2^2. \label{eq:prop_addition}
\end{equation}

\paragraph{Linear layer.}
If a vector $\mathbf{v} = (v_1, \ldots, v_D)$ of dimension $D$ containing elements of variance $\sigma_D^2$ is passed to a (zero-bias or biasless) linear layer of output dimension $T$, then given weights $w_{ij}$ of variance $\sigma_w^2$, the expected variance of the output elements $\mathbf{u} = (u_1, \ldots, u_T)$ is
\begin{align}
\sigma_T^2 &= \text{Var}(u_i) = \text{Var}\bigg(\sum_{j=1}^{D} w_{ij} v_j\bigg) = \sum_{i=1}^{D} \text{Var}(w_{ij} v_j) \notag \\
&= \sum_{i=1}^{D} \text{Var}(w_{ij}) \cdot \text{Var}(v_{j}) = D \cdot \sigma_w^2 \cdot \sigma_D^2. \label{eq:prop_linear}
\end{align}

\subsection{Variance-Balanced Transformer Initialization}
\label{app:weight_init_transformer}

In terms of the layers that are relevant for the statistical analysis of the data passing through the model during training, the object decoder fundamentally consists of (cf.~\Cref{fig:object_decoder}):
\begin{itemize}
    \item \textbf{Input stage:} A linear projection of the input embedding vector in parallel to the token embeddings of the ground truth object noun tokens, followed by an additive positional embedding,
    \item \textbf{Transformer:} $L$ consecutive transformer decoder layers, each sequentially consisting of two residual blocks, first for multi-head attention and then for feedforward layers, and,
    \item \textbf{Output stage:} A layer normalization, followed by a linear projection to obtain the token logits, where the weights of the projection are tied to the token embeddings matrix.
\end{itemize}
Note that no biases are used throughout the architecture, including in all linear, layer norm, and multi-head attention layers, and thus can be mathematically neglected in the analysis. If they were present, they would be initialized to zero anyway however, to allow them to be used in the long run by the optimizer, but initially not affect the optimization too much. We proceed with details of the variance-balanced weight initialization strategy for each part of the transformer model.

\paragraph{Input stage.}
The input to NOVIC is a unit embedding vector $\mathbf{e} = (e_1, \ldots, e_F)$ of dimension $F$ (\eg $F = 768$ for the nominal CLIP embedder), so the elements of this input signal vector have a zero-mean variance of
\begin{equation}
\sigma_e^2 = \frac{1}{F} \sum_{i=1}^{F} e_i^2 = \frac{1}{F} \| \mathbf{e} \|^2 = \frac{1}{F}. \label{eq:embedding_variance}
\end{equation}
To ensure approximate unit scale throughout the transformer post-initialization, we wish for each signal element throughout the network to have near-unit variance. As such, we initialize the weights $W_p$ of the linear projection layer from a zero-mean normal distribution of variance $\tfrac{1}{2}$, \ie $\mathcal{N}(0, \tfrac{1}{2})$, and do the same for all weights in the token embeddings $W_{te}$ and learned position embeddings $W_{pe}$. If $W_{te}^*$ is the learnable token embeddings for the specific ground truth tokens being passed to the object decoder, then from \Cref{eq:prop_addition,eq:prop_linear,eq:embedding_variance}, we can see that the expected variance $\sigma_p^2$ post-positional embedding, \ie of the actual input sequence vectors to the transformer layers, is thus
\begin{align}
\sigma_p^2 &= \text{Var} \Big( \text{Concat}(\mathbf{e} W_p, W_{te}^*) + W_{pe} \Big) \notag \\
&= \text{Concat} \Big( \text{Var}(\mathbf{e} W_p), \text{Var}(W_{te}^*) \Big) + \text{Var}(W_{pe}) \notag \\
&= \text{Concat} \Big( F \cdot \text{Var}(W_p) \cdot \sigma_e^2, \tfrac{1}{2} \Big) + \tfrac{1}{2} \notag \\
&= \text{Concat} \Big( F \cdot \tfrac{1}{2} \cdot \tfrac{1}{F}, \tfrac{1}{2} \Big) + \tfrac{1}{2} \notag \\
&= 1,
\end{align}
as desired. Note that some simplifying leniency is being used in the notation here by referring to elementwise variances with $\text{Var}(\cdot)$ of a matrix, and to work around the detail that the $PH$ outputs of the linear projection layer need to be split into $P$ (number of prefix tokens) individual sequence vectors of dimension $H$ for the dimensions to work out.

\paragraph{Transformer.}
The decoder-only transformer used in NOVIC consists of $L$ layers, each containing two residual blocks, for a total of $2L$ consecutive residual blocks that strictly add calculated residuals to the input signal but otherwise leave it untouched. Due to the random weights used in the calculation of the residual contributions, the residual blocks can be modeled as adding statistically independent noise to the main path that we want to ensure does not excessively dilute the input signal. We initialize the weights of each residual pathway so that it contributes an expected variance of $\tfrac{1}{2L}$, leading to a final expected output variance of $2$, of which half statistically comes directly from the input signal. The immediately following layer normalization of the output stage then renormalizes the scale of the data.

We achieve a residual pathway variance of $\tfrac{1}{2L}$ in each self-attention block by using weights of all 1 for the leading layer normalization, using weights of variance $\tfrac{1}{H}$ in order to make the query-key-value projections variance-preserving, and using weights of variance $\tfrac{1}{2LHS}$ in order to generate the desired output variance while accounting for (using $S$) the variance reduction that occurs in the scaled dot-product attention (SDPA). It can be shown that for $P$ prefix tokens and an expected input variance $\sigma^2$ to the SDPA, a good approximation of the variance reduction is given by the factor
\begin{equation}
S = \frac{1}{P} \bigg( 1 + \sigma^4 \frac{(P-1)}{P} \bigg).
\end{equation}
As the layer normalization layer outputs unit variance signals and the query-key-value projections are variance-preserving, the expected input variance to the SDPA is $\sigma^2 = 1$, so this reduces to
\begin{equation}
S = \frac{2P-1}{P^2}.
\end{equation}
For the residual pathway of each feedforward block we use a similar strategy, and use weights of all 1 for the leading layer normalization, use weights of variance $\tfrac{1}{H}$ in order to make the first feedforward linear layer variance-preserving (refer to \Cref{eq:prop_linear}), and use weights of variance $\tfrac{1}{2LKA}$ for the second feedforward linear layer, where $K$ is the intermediate feedforward dimension, and the factor $A$ compensates for the variance reduction of the activation function. $A$ is estimated by transforming the expected unit normal probability distribution by the activation function and calculating the variance around zero of the resulting distribution. For ReLU this yields $A = \tfrac{1}{2}$, and for GELU this yields $A \approx 0.4252$.

\paragraph{Output stage.}
As the weights of the token logits linear layer is tied to the already-initialized learnable token embeddings (variance $\tfrac{1}{2}$), it only remains to initialize the weight of the final layer normalization that precedes it (required due to the use of pre-layer normalization). As the final token logits need to be a suitable scale for the log-softmax operation that follows as part of the loss, we effectively `normalize' the output sequence vectors by initializing the output layer normalization weights to $\sigma_o = \tfrac{1}{\sqrt{H}}$. Using \Cref{eq:prop_linear}, this results in token logits of standard deviation
\begin{equation}
\sigma_l = \sqrt{H \cdot \tfrac{1}{2} \cdot \bigl(\tfrac{1}{\sqrt{H}}\bigr)^2} = \frac{1}{\sqrt{2}} \approx 0.707,
\end{equation}
which is a reasonable near-unit standard deviation that provides a good slightly conservative initialization of computed forward pass logits---crucially, without excessive input signal dilution---for fast stable convergence. The complete variance-balanced weight initialization strategy has been empirically confirmed at every layer of the model to work exactly like the statistical modeling suggests (also as a verification of the underlying independence assumptions), up to the expected approximations already discussed.

\section{Object Noun Dictionary}
\label{app:object_noun_dict}

The object noun dictionary was created based on a multitude of data sources, but most prominently the WordNet dictionary \cite{miller_wordnet_1995}, GNU Collaborative International Dictionary of English \cite{dict_gcide}, and the categories used by 35 different image classification and object detection/segmentation benchmarks. The benchmarks were divided into two groups, \emph{object} and \emph{general} benchmarks, depending on whether the class names were considered to be all valid object nouns or alternatively only used to provide frequency boosts, respectively. The steps used to generate the object noun dictionary were as follows:
\begin{enumerate}
\item All noun synsets in WordNet were collected that are part of one of the following 11 lexical classes (considered to be the only ones out of the 26 available that could \emph{possibly} contain object nouns): animal, artifact, body, event, feeling, food, object, person, phenomenon, plant, or possession.
\item The entire WordNet synset hierachy was formatted as a tree and used to manually select nodes and sub-hierarchies considered to fit the definition of an object noun. 1598 manual synset specifications were recorded and used to select \numprint{23942} synsets. This is the only step that required manual human intervention.
\item The chosen synsets were expanded to a working list of lemmas, and canonicalization was used to collect and deal with alternate spellings. A total of \numprint{42981} canonical lemmas resulted.
\item 4105 unique object names were loaded from the \emph{object} benchmarks and used to increase the number of canonical lemmas to \numprint{43588}.
\item A dataset of 1818 spelling equivalence mappings between American and British English were used to add additional spellings to the lemmas.
\item The GNU Collaborative International Dictionary of English was parsed and used to provide lemma noun alternates, and matches between singular and plural forms, especially for irregular nouns.
\item The WordNet plural noun exceptions list was used to deal with multiple singulars that share a plural form and multiple plurals that share a singular form.
\item A word inflection strategy was used to generate singular forms for nouns that had only appeared as plurals thus far, and the plural forms for all seen singulars.
\item The main spelling was resolved for each lemma with the use of language heuristics, and multiple lemmas that shared common spellings across singulars and plurals were merged. At this point there were \numprint{118045} noun alternates in total across the now \numprint{42919} lemmas.
\item The n-gram frequencies of the Google Web Trillion Word Corpus \cite{dict_ngrams,dict_trillion} were used to assign frequencies to each individual alternate of each lemma.
\item \numprint{68077} (non-unique) class names were loaded from the \emph{object} and \emph{general} benchmarks, and used to boost the frequencies of matching nouns in the dataset.
\item All noun frequencies were converted to an affine log-scale to better represent how often the corresponding nouns should be sampled during dataset generation.
\end{enumerate}

The final object noun dataset has \numprint{65210} singulars and \numprint{31316} plurals for a total of \numprint{96526} noun alternates that are associated with \numprint{42919} unique canonicalized object nouns. The average log-scale frequency of each noun alternate is 1.5, resulting in an average log-scale frequency of 3.5 per object noun.

\section{LLM-based Caption Generation}
\label{app:caption_gen}

\begin{promptfloat}
\begin{prompt}{caption_examples}{Singular/plural caption generation examples}
\raggedright
\psystem{You are a helpful AI assistant.} \\
\pprompt{Generate a list of 10 image captions each containing the target object ``bicycle'' (singular). Use the style of the image captions for large language model pre-training. Don't use poetic language. Use a different sentence structure for each caption. Output the list in the following form: \textless{}nr\textgreater{}. \textless{}caption\textgreater{}} \\
\presponse{\\
1. A red bicycle parked against a brick wall. \\
2. The bicycle resting under a tree in the park. \\
3. A woman riding a bicycle along the beach. \\
4. A vintage bicycle leaning against a lamppost. \\
5. Two children playing with a bicycle in the yard. \\
6. The bicycle locked to a metal railing outside the store. \\
7. A man carrying a folded bicycle up the stairs. \\
8. The bicycle lying on the grass with its wheels spinning. \\
9. A bicycle with a basket filled with colorful flowers. \\
10. The bicycle propped up against a wooden fence.
}
\tcbline
\raggedright
\psystem{You are a helpful AI assistant.} \\
\pprompt{Generate a list of 10 image captions each containing the target object ``bicycles'' (plural). Use the style of the image captions for large language model pre-training. Don't use poetic language. Use a different sentence structure for each caption. Output the list in the following form: \textless{}nr\textgreater{}. \textless{}caption\textgreater{}} \\
\presponse{\\
1. A row of bicycles parked by the street, waiting for their riders. \\
2. Bicycles lined up neatly in a bike-sharing station, ready for use. \\
3. Many bicycles secured to a rack outside a busy train station. \\
4. Several bicycles leaning against a wall, casting shadows in the afternoon sun. \\
5. Dozens of bicycles scattered across the open field during a cycling event. \\
6. A group of bicycles locked together in the urban bike parking area. \\
7. Bicycles parked in a row, their colorful frames catching the eye of passersby. \\
8. Multiple bicycles lined up near a city park, offering a convenient mode of transportation. \\
9. A cluster of bicycles parked haphazardly outside a bustling cafe. \\
10. Bicycles stacked on top of each other in a crowded bike storage area.
}
\end{prompt}
\end{promptfloat}

Given the complete object noun dictionary, an LLM was used to generate rich dynamic captions involving each of the object nouns in the dictionary. Based on the log-frequencies obtained from the object noun dictionary, ten times as many captions are generated independently for singular and plural forms of the noun, with at most 100 being generated for either category. OpenAI GPT-3.5 Turbo was used, specifically \emph{gpt-3.5-turbo-1106} with a temperature of 0.8, maximum 1000 output tokens, and example prompts/responses as shown in \autoref{prompt:caption_examples}. Ten captions are generated in every API call, and parsed using all available spelling variations of the object noun into appropriate caption templates. For instance, \emph{"Two children playing with a bicycle in the yard"} is parsed and stored as \emph{"Two children playing with a \{singular\} in the yard"}. In cases of words where there are multiple spellings, like for example \emph{orangutan}, this ensures that a caption template can be used in future to generate data for any desired spelling if required by the sampling strategy. A total of 1.8M captions were generated in this way.

\section{Effect of Noise Augmentation}
\label{app:noise_effect}

\begin{figure}[t]
    \parbox{\linewidth}{\centering%
    \includegraphics[width=\linewidth]{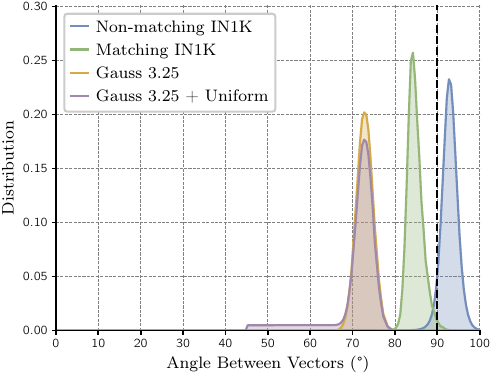}}
    \caption{\textbf{Comparison of angle separation distributions.} A plot of the distribution of embedding vector angle separations in 768-dimensional embedding space for matching and non-matching image-label pairs from the ImageNet-1K validation set. Also shown is the distribution of angle separations between original and noise-augmented text embeddings for both the pure Gaussian and Gaussian with uniform strategies. The dashed vertical line shows exactly perpendicular vectors that are thus mathematically uncorrelated.}
    \label{fig:modality_gap}
\end{figure}

Based on experiments, the `magnitude' of Gaussian noise that performed best for noise augmentation during training is $G = 3.25$. The precise definition of how this noise was implemented was by adding elementwise Gaussian noise with a standard deviation of
\begin{equation}
\sigma_G = \frac{G}{\sqrt{F}}
\end{equation}
to the input text embedding vectors and then renormalizing the result back to a unit vector. For the SigLIP B/16 model, which has an embedding dimension of $F = 768$, this corresponds to an elementwise standard deviation of $\sigma_G \approx 0.117$. If $\mathbf{g} = (g_1, \ldots, g_F)$ is the random noise vector with $g_i \sim \mathcal{N}(0, \sigma_G^2)$, then we can see that
\begin{align}
\mathbb{E}\bigl[\|\mathbf{g}\|^2\bigr] &= \mathbb{E}\biggl[ \sum_{i=1}^F g_i^2 \biggr] = \sum_{i=1}^F \mathbb{E}[g_i^2] = F \sigma_G^2 = G^2, \\
\text{Var}\bigl(\|\mathbf{g}\|^2\bigr) &= \text{Var} \biggl( \sum_{i=1}^F g_i^2 \biggr) = \sum_{i=1}^F \text{Var}(g_i^2) = \sum_{i=1}^F 2\sigma_G^4 \notag \\
&= 2F\Bigl(\frac{G}{\sqrt{F}}\Bigr)^4 = \frac{2G^4}{F}.
\end{align}
For our case, this means the random noise vector has an average square-norm $\|\mathbf{g}\|^2$ of $G^2 \approx 10.56$ with a standard deviation of $G^2\cdot\sqrt{\tfrac{2}{F}} \approx 0.539$. This clearly shows that statistically, the added noise vector will \emph{always} have a norm significantly larger (on the order of $\|\mathbf{g}\| \approx G = 3.25$) than the unit embedding vector $\mathbf{e} = (e_1, \ldots, e_F)$, which has $\|\mathbf{e}\| = 1$. On top of being such a large vector, due to the high dimensionality of the embedding space, the random noise vector $\mathbf{g}$ is actually also essentially \emph{guaranteed} to be near-perpendicular to the embedding vector $\mathbf{e}$, causing a large directional change of the embedding vector no matter what---no embedding vector can ever remain anywhere close to where it started. To demonstrate this, we consider the cosine of the angle $\theta$ between $\mathbf{e}$ and $\mathbf{g}$, given by
\begin{equation}
\cos{\theta} = \frac{\mathbf{e} \cdot \mathbf{g}}{\|\mathbf{e}\| \|\mathbf{g}\|} = \sum_{i=1}^{F} e_i \frac{g_i}{\|\mathbf{g}\|}.
\end{equation}
Due to the statistical independence of each element,
\begin{align}
\mathbb{E}[\cos{\theta}] &= \mathbb{E}\biggl[ \sum_{i=1}^{F} e_i \frac{g_i}{\|\mathbf{g}\|} \biggr] \notag \\
&= \sum_{i=1}^{F} e_i \cancelto{0}{\mathbb{E}\biggl [\frac{g_i}{\|\mathbf{g}\|} \biggr]} \notag \\
&= 0,
\end{align}
due to the symmetrical isotropic Gaussian nature of $\textbf{g}$. The corresponding variance is then
\begin{align}
\text{Var}(\cos{\theta}) &= \mathbb{E}[(\cos{\theta})^2] - \cancelto{0}{\mathbb{E}[\cos{\theta}]^2} \notag \\
&= \mathbb{E} \Biggl[ \biggl( \sum_{i=1}^{F} e_i \frac{g_i}{\|\mathbf{g}\|} \biggr)^2 \Biggr] \notag \\
&= \sum_{i=1}^{F} e_i^2 \, \mathbb{E} \biggl[ \frac{g_i^2}{\|\mathbf{g}\|^2} \biggr] + \sum_{i \neq j}^{F} e_i e_j \cancelto{0}{\mathbb{E} \biggl[ \frac{g_i g_j}{\|\mathbf{g}\|^2} \biggr]} \notag \\
&= \sum_{i=1}^{F} e_i^2 \cdot \frac{1}{F} \notag \\
&= \frac{1}{F},
\end{align}
as $\mathbf{e}$ is a unit vector. In other words, the cosine similarity between an embedding vector and the noise vector that is added to it is on average 0 (perpendicular) with a very narrow standard deviation of $\tfrac{1}{\sqrt{F}} \approx 0.036$, implying essentially guaranteed near-perpendicularity. The average angle $\beta$ between the initial embedding vector $\mathbf{e}$ and the noise augmented vector $\tfrac{\mathbf{e} + \mathbf{g}}{\|\mathbf{e} + \mathbf{g}\|}$ is thus approximately
\begin{equation}
\beta \approx \text{atan}\biggl( \frac{\|\mathbf{g}\|}{\|\mathbf{e}\|} \biggr) \approx \tan^{-1}{G} \approx 72.9\degree.
\end{equation}
The standard deviation of the angle $\beta$ can empirically be measured to be 1.94$\degree$. This narrow distribution of $\beta$, \ie the distribution of angle separations that occur when using the nominal Gaussian noise augmentation strategy, is shown in \Cref{fig:modality_gap}. Also shown in the plot is the measured distribution of angle separations for \emph{matching} and \emph{non-matching} image-label pairs of the ImageNet-1K validation set, \ie demonstrating the modality gap as well as the very thin separation that CLIP models provide between matching and non-matching pairs. The positive distribution-widening effect of the nominal strategy of adding 15\% uniform angle noise from 45--75$\degree$ is also shown (labeled ``Gauss 3.25 $+$ Uniform'' in \Cref{fig:modality_gap}). With this nominal strategy, for 85\% of all samples Gaussian noise of magnitude $G = 3.25$ is added, and for the remaining 15\% of samples (randomly chosen), uniform angle noise of random magnitude 45--75$\degree$ is applied.

\section{Open Vocabulary Image Datasets}
\label{app:open_vocab_datasets}

Three open vocabulary image datasets were collected as part of this work, World, Wiki and Val3K, as described in the following sections. All three datasets were annotated both manually by humans and by a multimodal LLM (cf.~\Cref{app:llm_annotation}),\footnote{All except human annotations for Val3K.} using one of five labels for each predicted object noun---\emph{correct primary} (1.0 points), \emph{correct secondary} (0.8 points), \emph{close primary} (0.5 points), \emph{close secondary} (0.4 points), and \emph{incorrect} (0 points). The sum of all points is divided by the number of images to get the final percent accuracy \emph{prediction score}. The suffixes -H and -L are used for the datasets to signify which annotations are being used, leading to five open vocabulary evaluation datasets, World-H, World-L, Wiki-H, Wiki-L, and Val3K-L.

\subsection{World Dataset}
\label{sec:world_dataset}

The World dataset was collected with the express intention of trying to curate a dense collection of both regular and tricky object concepts. This included a specific focus on trying to include images from around the world that have never been on the internet (and thus cannot ever have been trained by a CLIP model before). Tricky images include ones that show: unusual objects, common object concepts but in peculiar variations, deceptive objects that are one thing but styled to look somewhat like another, and indirect representations of objects, \eg paintings, cartoons, posters, sculptures, stuffed toys, drone show patterns, and such. A total of 12 people collected 85.3\% of the images from original photographs in 10 countries of the world (Australia, Austria, England, France, Germany, Netherlands, New Zealand, Philippines, UAE, USA), with an active focus on covering as wide and varied concepts as possible. The remaining 14.7\% of images were complemented from web sources to help cover concepts (\eg animals) that could not so readily be manually photographed. The 272 images consist of 15.4\% animals, 10.0\% plants, 16.2\% wider scenes, 6.6\% indirect representations of objects, and 51.8\% focused images of objects. Both human (World-H) and LLM (World-L) annotations are available for a wide variety of evaluated models.

\subsection{Wiki Dataset}
\label{sec:wiki_dataset}

The Wiki dataset was collected by scraping \numprint{18660} lead images, \ie `title images', from Wikipedia articles. Lead images often show a clear object concept that can be predicted and annotated, reducing the number of ambiguities and grey zones that need to be dealt with consistently during annotation. The scraped images had a noticeable bias in their topic of focus towards plants, animals, and people, so the sampling of the scraped images to a dataset size of \numprint{1000} was guided by a probability-adjusted sampling method to conservatively correct that bias. Both human (Wiki-H) and LLM (Wiki-L) annotations were made for a variety of evaluated models.

\subsection{Val3K Dataset}
\label{sec:val3k_dataset}

The Val3K dataset aims to help quantify how well a model is \emph{really} doing on the ImageNet-1K validation set, if it is not forced to output lower probability classifications in order to align with the class names of the dataset. The dataset was collected by sampling 3 random validation images (out of the 50 available) for each of the \numprint{1000} ImageNet-1K classes. This led to a dataset of \numprint{3000} images of mixed content. LLM annotations were made on all \numprint{3000} images to obtain Val3K-L. As LLM annotations, especially when passing images to a multimodal LLM, are attributed with cost, it was not feasible to perform annotations on the full \numprint{50000} ImageNet-1K validation set.

\section{Multimodal LLM Image Annotation}
\label{app:llm_annotation}

Due to the recent availability of some multimodal LLMs with a very high level of image understanding, it is conceivable to approximately quantify the performance of open vocabulary classification models by auto-annotating their predictions on image sets using these LLMs. This is particularly useful when it is just not feasible to perform the required amount of annotations as a human. The general process is as follows:
\begin{enumerate}
    \item The chosen to-be-evaluated model(s) are inferenced on the required dataset of images, and the predicted object nouns are recorded (saving the top-k predictions for each image is also possible).
    \item For each individual image, the deduplicated set of all predicted object nouns across all models is collected.
    \item Groups of up to 15 of these predicted object nouns are evaluated per call to the LLM, until every object noun has up to 5 opinions from the LLM whether it is correct or incorrect (present in the image or not).
    \item A ground truth annotation is established per-image per-predicted object noun based on the opinions that were given by the LLM.
\end{enumerate}

\begin{promptfloat}
\begin{prompt}{multimodal_scoring}{Multimodal LLM image classification}
\psystem{You are an AI assistant that has one and only one narrow task that you should strictly adhere to at all times. Given an image and an enumerated list of nouns by the user, you should first describe everything you see in the image in complete detail, and then provide an exactly matching enumerated list that for each noun provided by the user (explicitly repeat the noun) strictly classifies the noun into one of exactly two categories using a single word - Correct or Incorrect. Correct means that at least one instance of the noun is visible in the image. Incorrect means nothing very visually similar to the noun is visible in the image.} \\
\pprompt{\emph{\textless{}image base64\textgreater{}} (max 512$\times$512)} \\
\parbox[c]{\dimexpr\textwidth-4cm\relax}{%
1) african hunting dog \\
2) african wild dog \\
3) canis niger \\
4) cape hunting dog \\
5) hyena \\
6) hyena dog \\
7) lycaon pictus \\
8) mammal \\
9) mammalian \\
10) mexican hairless \\
11) mexican hairless dog \\
12) monkey dog \\
13) okapi
}%
\parbox[c]{4cm}{%
\hfill\includegraphics[width=3.2cm]{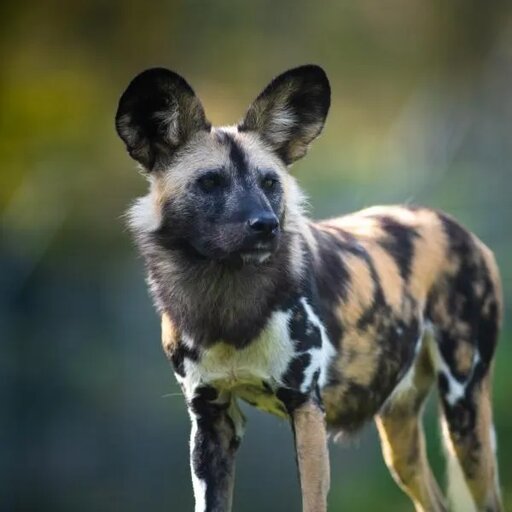}
} \\
\tcbline
\presponse{\\
The image shows an African wild dog, also known as Lycaon pictus. It has a distinctive coat with patches of black, brown, white, and yellow, large rounded ears, and a slender body. \\
1) african hunting dog - Correct \\
2) african wild dog - Correct \\
3) canis niger - Incorrect \\
4) cape hunting dog - Correct \\
5) hyena - Incorrect \\
6) hyena dog - Incorrect \\
7) lycaon pictus - Correct \\
8) mammal - Correct \\
9) mammalian - Correct \\
10) mexican hairless - Incorrect \\
11) mexican hairless dog - Incorrect \\
12) monkey dog - Incorrect \\
13) okapi - Incorrect
}
\end{prompt}
\vspace*{-2ex}
\end{promptfloat}

We use OpenAI GPT-4o, specifically \emph{gpt-4o-2024-05-13} with a temperature of 0.2, maximum 512 output tokens, and a cumulative probability threshold of 0.6 for nucleus sampling. The chosen parameters reflect the stability and consistency that is desired from the LLM. The prompt used for auto-annotation is shown in \autoref{prompt:multimodal_scoring} for an example image. Asking the LLM to first provide a description of the image allows it to adjust its summarized knowledge of the image according to the provided object noun candidates, and enhances its classification precision. For best results, 8--15 object noun candidates are always given, as providing too few or too many has a small potential of causing inconsistent annotations for fine-grained object nouns. With intentional randomization of candidate ordering and which candidate is picked for which API request, each object noun is annotated 3--5 times, stopping before 5 if there is at least 80\% agreement in the LLM's opinions. If the final agreement is at least 80\%, then the corresponding opinion (\emph{correct} or \emph{incorrect}) is taken as the final annotation, otherwise the annotation is \emph{close}. No \emph{primary} \vs \emph{secondary} classification is made because it was experimentally deemed too inconsistent. Due to the relatively unlikely condition required for the \emph{close} category (LLMs tend to be wrong more often than they are inconsistent at a low temperature), it is systematically not seen as often as for human annotations.

\begin{figure}[t]
    \parbox{\linewidth}{\centering%
    \includegraphics[width=\linewidth]{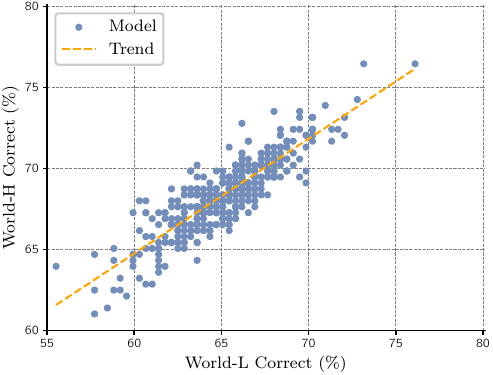}\\
    \includegraphics[width=\linewidth]{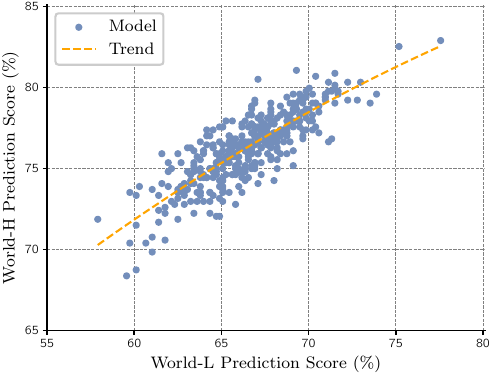}}
    \caption{\textbf{Comparison of human and LLM image annotations.} World-H scores are plotted against World-L scores for 397 models. \textbf{Top:} Scores considering \emph{correct} only. \textbf{Bottom:} Standard prediction scores considering \emph{close} classifications as well. For human annotations, \emph{secondary} classifications are scored as \emph{primary} as the LLM does not make that distinction.}
    \label{fig:human_vs_gpt}
\end{figure}

\begin{table}[t]
    \centering\small
    \setlength\tabcolsep{1.3pt}
    \begin{tabular}{lcc}
        \toprule
        Hyperparameter & Value & Tested Range \\
        \midrule
        CLIP model & SigLIP B/16 & See \Cref{tab:ablations} \\
        CLIP mixed precision & float16 & float16 \\
        Noun frequency threshold\hspace*{4.7pt} & 0 & 0--10 \\
        Multiset order $m$ & 3 & 1--4 \\
        LLM captions & \cmark & \xmark/\cmark \\
        Noise augmentation & Mix & Gauss, Uniform, Mix \\
        Gauss noise magnitude & 3.25 & 0.0--4.0 \\
        Uniform angle noise & 45--75 & Up to 0--85 \\
        Noise mix factor & 15\% & 0--20\% \\
        \midrule
        Prefix tokens $P$ & 4 & 2--6 \\
        Hidden dim $H$ & 512 & 384, 512 \\
        Feedforward dim $K$ & 128 & 64--512 \\
        Num layers $L$ & 6 & 4--12 \\
        Num heads & 8 & 8 \\
        Dropout & 0.1 & 0.05--0.2 \\
        Bias parameters & \xmark & \xmark/\cmark \\
        Label smoothing & \xmark & \xmark/\cmark \\
        Beam width & 10 & 1--10 \\
        Temperature & 1 & 0.33--3 \\
        \midrule
        Epochs $E$ & 18 & 9--24 \\
        Batch size $B$ & 8192 & 512--16384 \\
        Optimizer & AdamW & AdamW, AdamP \\
        Adam $\beta_1$ & 0.9 & 0.9 \\
        Adam $\beta_2$ & 0.95 & 0.95, 0.99 \\
        LR scheduler & cosine & cosine \\
        LR warmup epochs & 0 & 0--2 \\
        Initial LR & 1.5e-3 & 5e-4\,--\,3e-3 \\
        Weight decay & 0.1 & 0--0.3 \\
        Weight decay for 1D & \xmark & \xmark/\cmark \\
        \bottomrule
    \end{tabular}
    \caption{\textbf{NOVIC hyperparameter values and tested ranges.} Batch sizes use gradient accumulation up to factor 16.}
    \label{tab:hyperparameters}
\end{table}

In order to verify that the LLM annotations are sufficiently accurate to make inferences about the performance of evaluated NOVIC models, we compared the prediction scores computed on human and LLM annotations for 397 models on the World dataset (cf.~\Cref{fig:human_vs_gpt}). As expected, a clear correlation and trend can be identified, whether \emph{close} classifications are considered in the scores or not, with residual root mean square errors of 1.1\% in both cases. This suggests that even though the absolute values of the scores do noticeably differ, the coarse ordering it applies to models is similar.

\section{Model Hyperparameters}
\label{app:hyperparameters}

The nominal values of various NOVIC hyperparameters are shown in \Cref{tab:hyperparameters}. The corresponding hyperparameter value ranges that were covered in some configurations during testing are also shown.

\section{Qualitative Comparison}
\label{app:qualitative_comparison}

\afterpage{\clearpage%
\begin{table}[t]
    \centering\small
    \setlength\tabcolsep{1.3pt}
    \begin{tabular}{l@{\hspace{0.8em}}>{\centering\arraybackslash}p{1.12cm}>{\centering\arraybackslash}p{1.12cm}>{\centering\arraybackslash}p{1.12cm}}
        \toprule
        Configuration & World-H & Wiki-L & IN1K \\
        \midrule
        Prefix tokens $P = 2$ & 77.91 &  62.75 & 65.51 \\
        Prefix tokens $P = 4$* & \textbf{78.92} & 63.97 & \textbf{68.11} \\
        Prefix tokens $P = 6$ & 78.03 & \textbf{64.15} & 68.08 \\
        \midrule
        Layers $L = 4$ & 77.23 & 62.38 & 63.85 \\
        Layers $L = 6$* & \textbf{78.92} & 63.97 & 68.11 \\
        Layers $L = 8$ & 77.94 & 62.78 & 69.05 \\
        Layers $L = 12$ & 78.15 & \textbf{64.00} & \textbf{71.19} \\
        \midrule
        Hidden dim $H = 384$ & 76.97 & 61.65 & 65.77 \\
        Hidden dim $H = 512$* & \textbf{78.92} & \textbf{63.97} & \textbf{68.11} \\
        \midrule
        Batch size $B = 4096$ & 78.14 & 63.73 & 66.76 \\
        Batch size $B = 8192$* & 78.92 & \textbf{63.97} & \textbf{68.11} \\
        Batch size $B = 16384$ & \textbf{79.56} & 63.17 & 67.85 \\
        \midrule
        Epochs $E = 12$ & 77.33 & 63.30 & 66.33 \\
        Epochs $E = 18$* & \textbf{78.92} & \textbf{63.97} & \textbf{68.11} \\
        Epochs $E = 24$ & 77.77 & 63.00 & 67.71 \\
        \bottomrule
    \end{tabular}
    \caption{\textbf{Object decoder architecture and training hyperparameter ablations (\% mean of 3).} IN1K is the ImageNet-1K classification performance of the trained decoders, and the asterisks correspond to the nominal ablation baseline (FT2). World-H and Wiki-L are prediction scores (described in \Cref{sec:experiments}) on the corresponding datasets. See \Cref{sec:ablations} for more ablation results.}
    \label{tab:decoder_ablations}
\end{table}%
}

\begin{figure*}[p]
    \newlength{\imgcolwidth}%
    \setlength{\imgcolwidth}{3.6cm}%
    \newlength{\imgdim}%
    \setlength{\imgdim}{3.45cm}%
    \newlength{\imgrowsep}%
    \setlength{\imgrowsep}{0.6cm}%
    \parbox{\linewidth}{\small\centering%
    \parbox[t]{2.9cm}{\raggedleft\hfill\vspace*{\imgdim}\\Tag2Text\\RAM\\RAM (FT0)\\SigLIP NOVIC (FT2)\\SigLIP NOVIC (FT0)\\DFN-5B NOVIC (FT0)}\hspace*{0.15cm}%
    \parbox[t]{\imgcolwidth}{\centering\vspace*{0pt}\includegraphics[width=\imgdim,height=\imgdim]{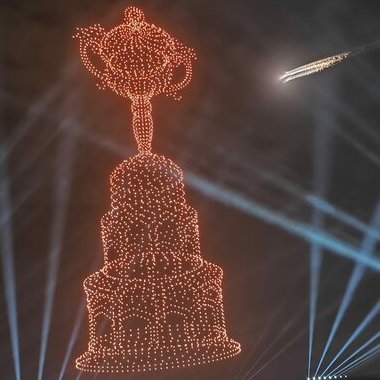}\\\red{stage (I)}\\\red{stage (I)}\\\red{stage (I)}\\\green{trophy cup (P)}\\\green{trophy cup (P)}\\firework (S)}%
    \parbox[t]{\imgcolwidth}{\centering\vspace*{0pt}\includegraphics[width=\imgdim,height=\imgdim]{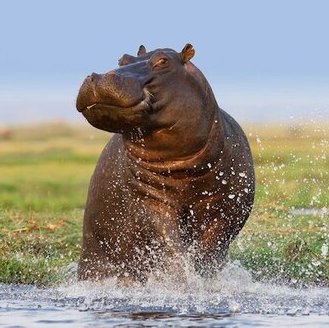}\\water (S)\\water (S)\\water (S)\\\green{hippopotamus amphibius (P)}\\\green{hippopotamus amphibius (P)}\\\green{hippopotamus (P)}}%
    \parbox[t]{\imgcolwidth}{\centering\vspace*{0pt}\includegraphics[width=\imgdim,height=\imgdim]{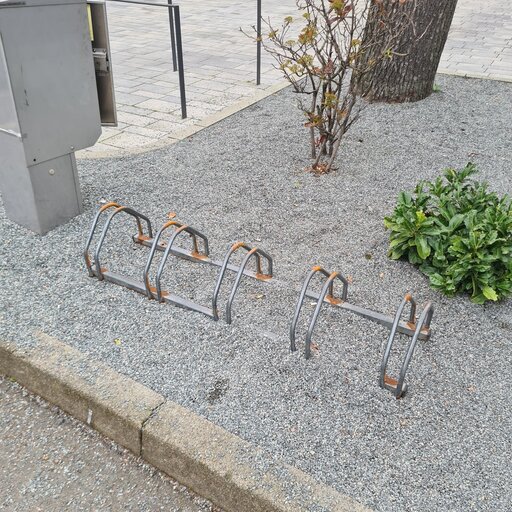}\\\red{ladder (I)}\\pavement (S)\\pavement (S)\\\green{bicycle rack (P)}\\\green{bicycle rack (P)}\\\green{bicycle rack (P)}}%
    \parbox[t]{\imgcolwidth}{\centering\vspace*{0pt}\includegraphics[width=\imgdim,height=\imgdim]{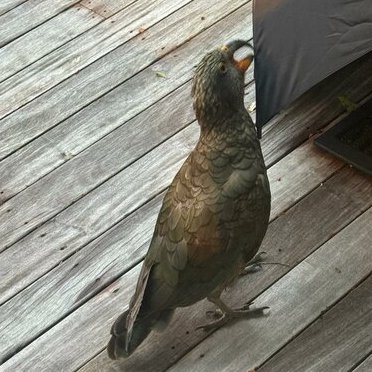}\\\green{bird (P)}\\\green{bird (P)}\\\green{bird (P)}\\parrot (CP)\\\red{mourning dove (I)}\\\green{kea (P)}}%
    \vspace*{\imgrowsep}
    \parbox[t]{2.9cm}{\raggedleft\hfill\vspace*{\imgdim}\\Tag2Text\\RAM\\RAM (FT0)\\SigLIP NOVIC (FT2)\\SigLIP NOVIC (FT0)\\DFN-5B NOVIC (FT0)}\hspace*{0.15cm}%
    \parbox[t]{\imgcolwidth}{\centering\vspace*{0pt}\includegraphics[width=\imgdim,height=\imgdim]{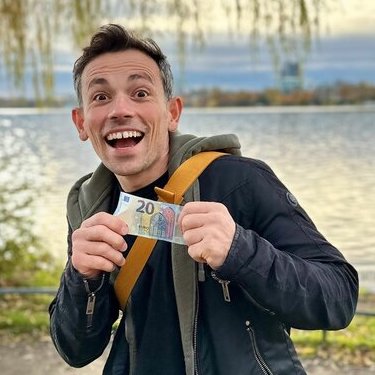}\\\green{man (P)}\\\green{man (P)}\\\green{man (P)}\\\green{money (P)}\\\green{money (P)}\\\green{banknote (P)}}%
    \parbox[t]{\imgcolwidth}{\centering\vspace*{0pt}\includegraphics[width=\imgdim,height=\imgdim]{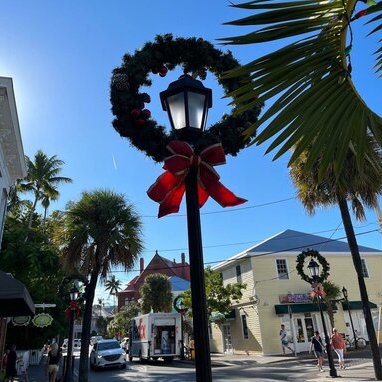}\\palm tree (S)\\palm tree (S)\\palm tree (S)\\\green{wreath (P)}\\\green{wreath (P)}\\\green{florida key (P)}}%
    \parbox[t]{\imgcolwidth}{\centering\vspace*{0pt}\includegraphics[width=\imgdim,height=\imgdim]{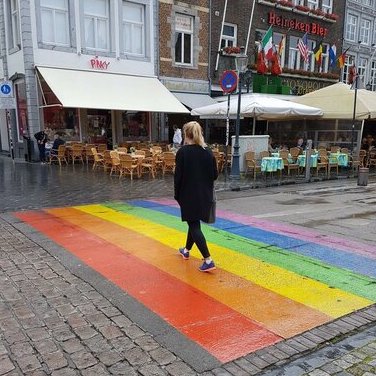}\\\red{walk (I)}\\\red{walk (I)}\\\green{female person (P)}\\\green{pavement (P)}\\\green{pedestrian crossing (P)}\\\green{pedestrian (P)}}%
    \parbox[t]{\imgcolwidth}{\centering\vspace*{0pt}\includegraphics[width=\imgdim,height=\imgdim]{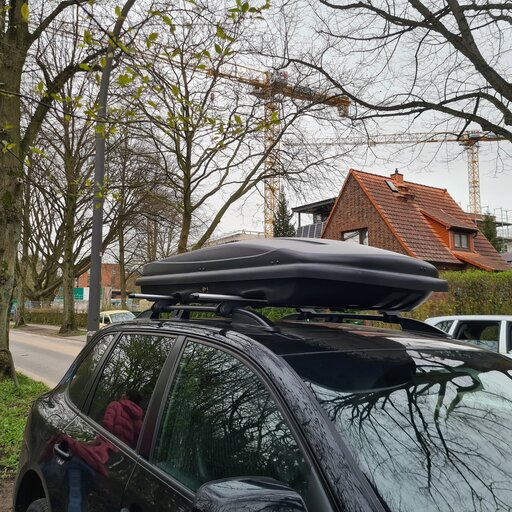}\\\green{car (P)}\\\green{car (P)}\\\green{car (P)}\\\green{roof rack (P)}\\\green{roof rack (P)}\\\green{roof rack (P)}}%
    \vspace*{\imgrowsep}
    \parbox[t]{2.9cm}{\raggedleft\hfill\vspace*{\imgdim}\\Tag2Text\\RAM\\RAM (FT0)\\SigLIP NOVIC (FT2)\\SigLIP NOVIC (FT0)\\DFN-5B NOVIC (FT0)}\hspace*{0.15cm}%
    \parbox[t]{\imgcolwidth}{\centering\vspace*{0pt}\includegraphics[width=\imgdim,height=\imgdim]{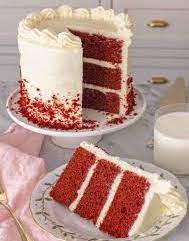}\\\green{cake (P)}\\\green{cake (P)}\\\green{cake (P)}\\\green{red velvet cake (P)}\\\red{cake cake (I)}\\\green{red velvet cake (P)}}%
    \parbox[t]{\imgcolwidth}{\centering\vspace*{0pt}\includegraphics[width=\imgdim,height=\imgdim]{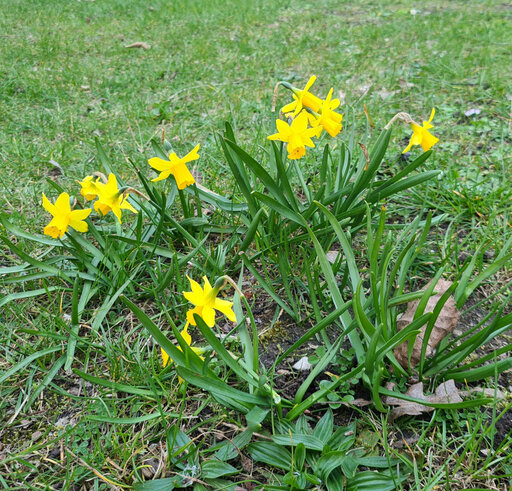}\\\green{flower (P)}\\\red{yellow (I)}\\\green{flower (P)}\\\green{jonquil (P)}\\\green{jonquil (P)}\\\green{daffodil (P)}}%
    \parbox[t]{\imgcolwidth}{\centering\vspace*{0pt}\includegraphics[width=\imgdim,height=\imgdim]{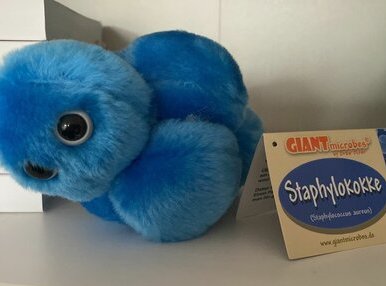}\\\red{animal (I)}\\\red{blue (I)}\\\red{blue (I)}\\\green{stuffed toy (P)}\\\green{stuffed toy (P)}\\\green{genus staphylococcus (P)}}%
    \parbox[t]{\imgcolwidth}{\centering\vspace*{0pt}\includegraphics[width=\imgdim,height=\imgdim]{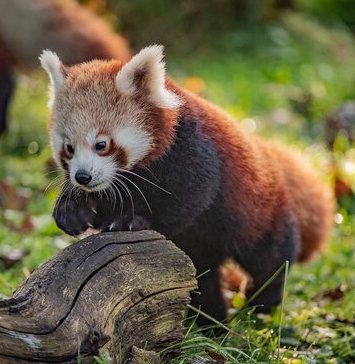}\\\green{panda (P)}\\\red{red (I)}\\\red{red cole (I)}\\\green{red panda (P)}\\\green{red panda (P)}\\\green{ailurus (P)}}%
    \vspace*{0.3cm}
    }
    \caption{\textbf{Examples of open vocabulary classification on the World-H dataset.} While RAM and Tag2Text select the most likely noun from a limited vocabulary, NOVIC can freely output any object noun. Possible classification scores are \emph{correct primary} (P), \emph{correct secondary} (S), \emph{close primary} (CP), \emph{close secondary} (CS), and \emph{incorrect} (I). \green{Correct primary} and \red{incorrect} labels are highlighted in color. Output labels that are not nouns are uninformative for the image and thus considered incorrect. Similarly, invalid nouns are also deemed incorrect.}
    \label{fig:qualitative-world}
\end{figure*}

\begin{figure*}[p]
    \setlength{\imgcolwidth}{3.6cm}%
    \setlength{\imgdim}{3.45cm}%
    \setlength{\imgrowsep}{0.6cm}%
    \parbox{\linewidth}{\small\centering%
    \parbox[t]{2.9cm}{\raggedleft\hfill\vspace*{\imgdim}\\Tag2Text\\RAM\\RAM (FT0)\\SigLIP NOVIC (FT2)\\SigLIP NOVIC (FT0)\\DFN-5B NOVIC (FT0)}\hspace*{0.15cm}%
    \parbox[t]{\imgcolwidth}{\centering\vspace*{0pt}\includegraphics[width=\imgdim,height=\imgdim]{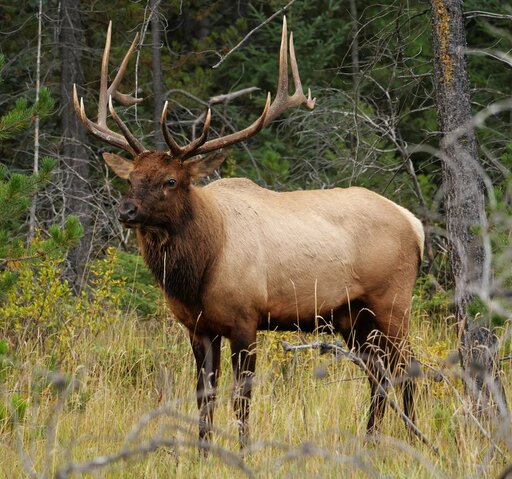}\\\green{elk (P)}\\moose (CP)\\\red{stand (I)}\\\red{buckskin (I)}\\\red{moose wood (I)}\\\green{elk (P)}}%
    \parbox[t]{\imgcolwidth}{\centering\vspace*{0pt}\includegraphics[width=\imgdim,height=\imgdim]{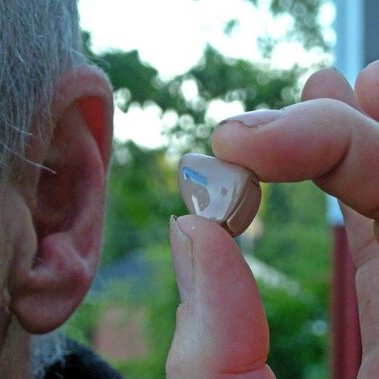}\\man (S)\\hand (S)\\finger (S)\\\green{hearing aid (P)}\\\green{hearing aid (P)}\\\green{hearing aid (P)}}%
    \parbox[t]{\imgcolwidth}{\centering\vspace*{0pt}\includegraphics[width=\imgdim,height=\imgdim]{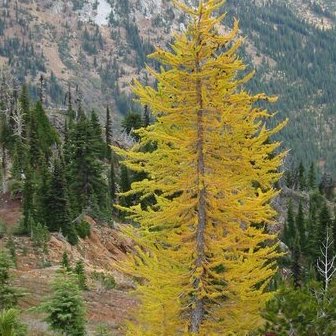}\\\green{tree (P)}\\\green{tree (P)}\\\green{tree (P)}\\brewers spruce (CP)\\western larch (CP)\\\green{subalpine larch (P)}}%
    \parbox[t]{\imgcolwidth}{\centering\vspace*{0pt}\includegraphics[width=\imgdim,height=\imgdim]{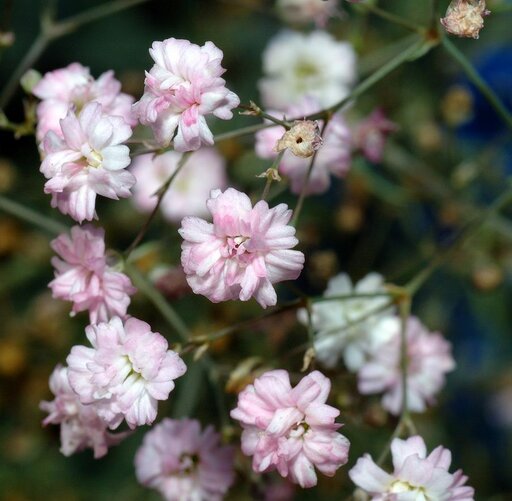}\\\green{flower (P)}\\\green{flower (P)}\\\green{flower (P)}\\\green{babies breath (P)}\\genus saponaria (CP)\\\green{genus gysophila (P)}}%
    \vspace*{\imgrowsep}
    \parbox[t]{2.9cm}{\raggedleft\hfill\vspace*{\imgdim}\\Tag2Text\\RAM\\RAM (FT0)\\SigLIP NOVIC (FT2)\\SigLIP NOVIC (FT0)\\DFN-5B NOVIC (FT0)}\hspace*{0.15cm}%
    \parbox[t]{\imgcolwidth}{\centering\vspace*{0pt}\includegraphics[width=\imgdim,height=\imgdim]{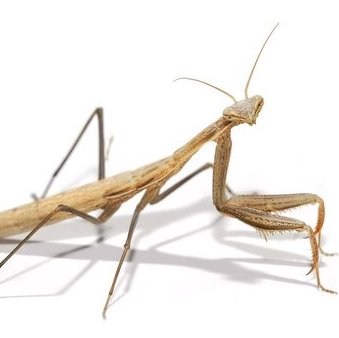}\\\green{mantid (P)}\\\red{photo (I)}\\\green{mantis (P)}\\\green{praying mantis (P)}\\\green{praying mantis (P)}\\\green{mantis (P)}}%
    \parbox[t]{\imgcolwidth}{\centering\vspace*{0pt}\includegraphics[width=\imgdim,height=\imgdim]{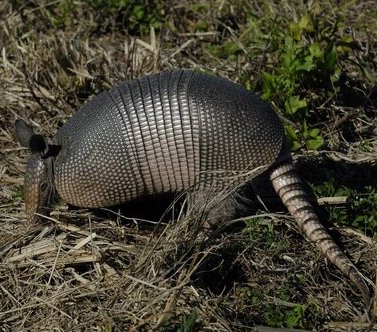}\\grass (CS)\\\green{armadillo (P)}\\\green{armadillo (P)}\\\green{nine banded armadillo (P)}\\\green{armadillo (P)}\\\green{dasypus (P)}}%
    \parbox[t]{\imgcolwidth}{\centering\vspace*{0pt}\includegraphics[width=\imgdim,height=\imgdim]{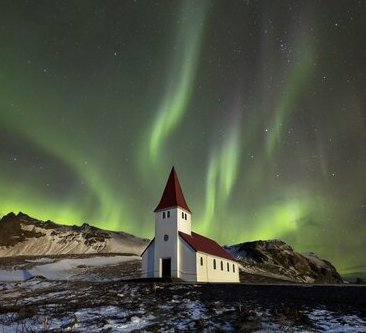}\\\green{church (P)}\\\green{church (P)}\\\green{church (P)}\\\green{church (P)}\\\green{church (P)}\\\green{aurora (P)}}%
    \parbox[t]{\imgcolwidth}{\centering\vspace*{0pt}\includegraphics[width=\imgdim,height=\imgdim]{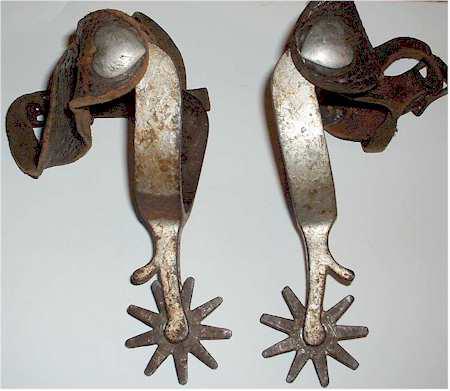}\\\red{old (I)}\\\green{tool (P)}\\\green{tool (P)}\\\green{spur (P)}\\\green{spur (P)}\\\green{spur (P)}}%
    \vspace*{\imgrowsep}
    \parbox[t]{2.9cm}{\raggedleft\hfill\vspace*{\imgdim}\\Tag2Text\\RAM\\RAM (FT0)\\SigLIP NOVIC (FT2)\\SigLIP NOVIC (FT0)\\DFN-5B NOVIC (FT0)}\hspace*{0.15cm}%
    \parbox[t]{\imgcolwidth}{\centering\vspace*{0pt}\includegraphics[width=\imgdim,height=\imgdim]{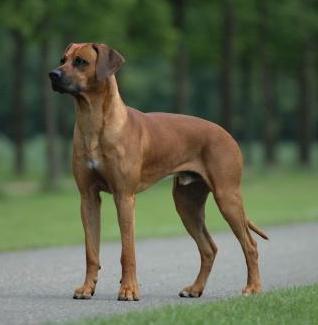}\\\green{dog (P)}\\\green{dog (P)}\\\green{dog (P)}\\\green{rhodesian ridgeback (P)}\\\green{rhodesian ridgeback (P)}\\\green{rhodesian ridgeback (P)}}%
    \parbox[t]{\imgcolwidth}{\centering\vspace*{0pt}\includegraphics[width=\imgdim,height=\imgdim]{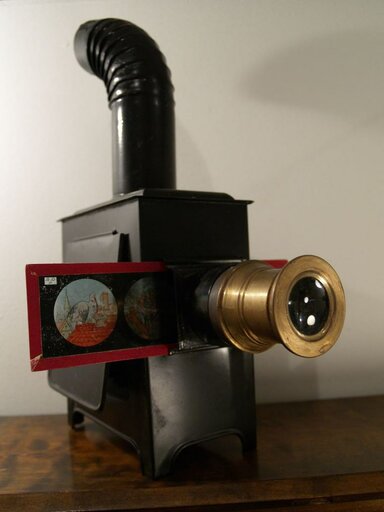}\\telescope (CP)\\camera (CP)\\camera (CP)\\\red{viewfinder (I)}\\\red{viewfinder (I)}\\\green{magic lantern (P)}}%
    \parbox[t]{\imgcolwidth}{\centering\vspace*{0pt}\includegraphics[width=\imgdim,height=\imgdim]{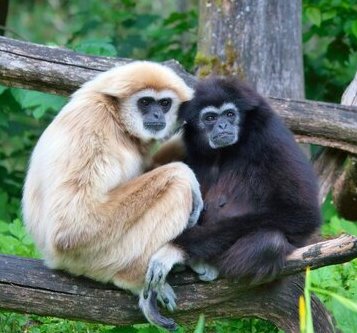}\\\green{monkey (P)}\\\red{sit (I)}\\\green{monkey (P)}\\\green{gibbon (P)}\\hylobates syndactylus (CP)\\\green{gibbon (P)}}%
    \parbox[t]{\imgcolwidth}{\centering\vspace*{0pt}\includegraphics[width=\imgdim,height=\imgdim]{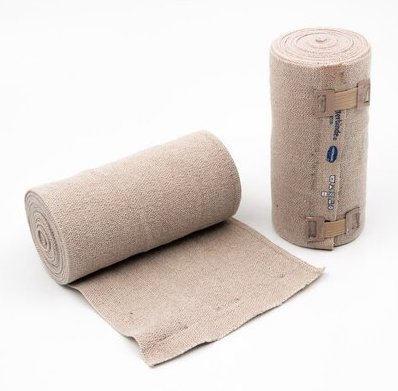}\\\green{roll (P)}\\\red{roll on (I)}\\\green{roll (P)}\\\green{plaster bandage (P)}\\\green{compression bandage (P)}\\\green{compression bandage (P)}}%
    \vspace*{0.3cm}
    }
    \caption{\textbf{Examples for open vocabulary classification on the Wiki-H dataset.} The images are an assortment of Wikipedia lead images, covering a broad range of visual concepts. Possible classification scores are \emph{correct primary} (P), \emph{correct secondary} (S), \emph{close primary} (CP), \emph{close secondary} (CS), and \emph{incorrect} (I). \green{Correct primary} and \red{incorrect} labels are highlighted in color.}
    \label{fig:qualitative-wiki}
\end{figure*}

A qualitative comparison between NOVIC and image tagging baselines on the World-H and Wiki-H datasets is presented in \Cref{fig:qualitative-world} and \Cref{fig:qualitative-wiki}, respectively. RAM and Tag2Text in general provide relatively coarse-grained classifications and often miss central concepts in the images, focusing instead on known nouns. For example, they output \emph{man} for an image where a man is presenting a \emph{banknote} to the camera. Additionally, they sometimes output colors (\eg \emph{blue}, \emph{yellow}), verbs (\eg \emph{walk}, \emph{stand}), attributes (\eg \emph{old}), or merely image style information (\eg \emph{photo}), thereby failing to convey the actual content of the image. Even when given access to the complete object noun dictionary (\ie FT0), which is an order of magnitude more comprehensive than the RAM vocabulary, RAM predominantly predicts the familiar words it was trained on rather than utilizing a broad range of object nouns as labels.

On the other hand, NOVIC demonstrates strong capabilities in precisely detecting and outputting the central object of an image. Particularly when trained on the DFN-5B CLIP model, NOVIC provides very fine-grained classifications and can accurately identify Latin names of plants and animals (\eg \emph{genus gypsophila}), names of very specific uncommon objects (\eg \emph{magic lantern}), and even geolocations of pictures (\eg \emph{florida key}) with high precision. Since NOVIC generates free-form text during inference, it occasionally fails to ``stop in time'' when outputting nouns however, resulting in incorrect labels (\eg \emph{cake cake}, \emph{moose wood}). This issue is more prevalent for smaller CLIP models and lower word frequency thresholds of the training dataset.

\section{Decoder Architecture Ablations}
\label{app:decoder_arch_ablations}

Ablations of the core NOVIC model hyperparameters are shown in \Cref{tab:decoder_ablations}. The results support the nominal choice of hyperparameters, and suggest that slight gains could potentially be made with a greater number of transformer layers (although model generalization becomes slightly less reliable) and/or batch size. Refer to \Cref{sec:ablations} for ablation results pertaining to the training dataset, noise augmentation, and CLIP model.

\section{Image Classification Benchmark Trends}
\label{app:cls_perf_discussion}

\Cref{tab:image_cls} shows the performance of NOVIC on various image classification benchmarks, many of which are commonly used to benchmark the zero-shot image classification performance of CLIP models. In general, the trend can be seen that the classification performance of NOVIC models slowly drops off as the complexity and extensiveness of the training data is increased towards the full size of the object noun dictionary. This can be expected, as the object decoder has increasingly many object concepts to learn as the frequency threshold (FT) decreases, and these concepts are also increasingly densely packed in the embedding space, making them harder to accurately keep apart. An exception to this general trend for some of the classification benchmarks is the FT9 score, as the FT9 training data does not actually always cover all of the target nouns required for each dataset. This concretely means that $412/1000$ ImageNet-1K classes, $17/200$ Tiny ImageNet classes, $50/200$ ImageNet-A classes, and $37/200$ ImageNet-R classes, cannot contribute to the final classification score, resulting in the disadvantaged lower scores marked with a dagger ($\dagger$) in \Cref{tab:image_cls}. Another exception to this general trend are the benchmarks that have few and coarse-grained classes, such as Imagenette and CIFAR-10. For these datasets, the performance depends less on the diversity of object nouns seen during training, and more on just the raw ability of the CLIP image encoder to produce an accurate and representative embedding, resulting in fairly consistent classification performances irrespective of the frequency threshold.

\begin{table*}[t]
    \centering\small
    \setlength\tabcolsep{1.3pt}
    \begin{tabular}{p{1.9cm}>{\centering\arraybackslash}p{1.8cm}>{\centering\arraybackslash}p{1.3cm}>{\centering\arraybackslash}p{1.3cm}>{\centering\arraybackslash}p{1.3cm}>{\centering\arraybackslash}p{1.1cm}>{\centering\arraybackslash}p{1.3cm}c>{\centering\arraybackslash}p{1.3cm}c>{\centering\arraybackslash}p{1.2cm}}
        \toprule
        Model & Prompt-free Classification & Text-only Training & Seen Objects & Used Objects & RAM Vocab & Primary Score && \hspace*{-0.1cm}\parbox[t]{1.5cm}{\centering{}Prediction Score}\hspace*{-0.1cm} && Overall Score \\
        \midrule
        Tag2Text \cite{huang_tag2text_2024} & \xmark & \xmark & \numprint{3429} & 176 & 91.9 & 62.32 & ${\scriptstyle\xleftarrow{\textcolor{red}{-15.29}}}$ & 77.61 & ${\scriptstyle\xrightarrow{\textcolor{red}{-13.49}}}$ & 64.12 \\
        RAM \cite{zhang_recognize_2024} & \xmark & \xmark & \numprint{4585} & 160 & 100.0 & 57.35 & ${\scriptstyle\xleftarrow{\textcolor{red}{-17.65}}}$ & 75.00 & ${\scriptstyle\xrightarrow{\textcolor{red}{-15.86}}}$ & 59.14 \\
        RAM (FT2) & \xmark & \xmark & \numprint{11897} & 188 & 76.8 & 64.71 & ${\scriptstyle\xleftarrow{\textcolor{red}{-11.61}}}$ & 76.32 & ${\scriptstyle\xrightarrow{\textcolor{red}{-13.84}}}$ & 62.48 \\
        RAM (FT0) & \xmark & \xmark & \textbf{\numprint{42919}} & 188 & 74.3 & 63.97 & ${\scriptstyle\xleftarrow{\textcolor{red}{-11.03}}}$ & 75.00 & ${\scriptstyle\xrightarrow{\textcolor{red}{-13.84}}}$ & 61.16 \\
        NOVIC (FT9) & \cmark & \cmark & \numprint{2919} & 235 & 66.9 & 74.82 & ${\scriptstyle\xleftarrow{\textcolor{Peach}{-2.79}}}$ & 77.61 & ${\scriptstyle\xrightarrow{\textcolor{Peach}{-3.22}}}$ & 74.39 \\
        NOVIC (FT6) & \cmark & \cmark & \numprint{5899} & 252 & 58.5 & 76.47 & ${\scriptstyle\xleftarrow{\textcolor{Peach}{-2.21}}}$ & 78.68 & ${\scriptstyle\xrightarrow{\textcolor{Green}{-1.01}}}$ & 77.67 \\
        NOVIC (FT2) & \cmark & \cmark & \numprint{11897} & 253 & 52.6 & 77.94 & ${\scriptstyle\xleftarrow{\textcolor{Green}{-1.62}}}$ & 79.56 & ${\scriptstyle\xrightarrow{\textcolor{Green}{-0.64}}}$ & 78.92 \\
        NOVIC (FT0) & \cmark & \cmark & \textbf{\numprint{42919}} & 251 & 49.6 & 71.14 & ${\scriptstyle\xleftarrow{\textcolor{Peach}{-3.53}}}$ & 74.67 & ${\scriptstyle\xrightarrow{\textcolor{Green}{-1.10}}}$ & 73.57 \\
        NOVIC${}^\dagger$ (FT2) & \cmark & \cmark & \numprint{11897} & \textbf{263} & 55.5 & 86.77 & ${\scriptstyle\xleftarrow{\textcolor{Green}{-1.17}}}$ & 87.94 & ${\scriptstyle\xrightarrow{\textcolor{Green}{-0.81}}}$ & 87.13 \\
        NOVIC${}^\dagger$ (FT0) & \cmark & \cmark & \textbf{\numprint{42919}} & 260 & \textbf{48.5} & \textbf{86.95} & ${\scriptstyle\xleftarrow{\textcolor{Green}{-1.32}}}$ & \textbf{88.27} & ${\scriptstyle\xrightarrow{\textcolor{Green}{-0.37}}}$ & \textbf{87.90} \\
        \bottomrule
    \end{tabular}
    \caption{\textbf{Open vocabulary World-H results in comparison to baselines (best of 3).} Tag2Text and RAM achieve artificially high scores by frequently outputting uninformative generic tags like \emph{man}, \emph{tree}, \emph{grass}, \emph{plate}, and often miss the actual intent of an image. Scores are in \%. \emph{RAM Vocab} is the percent of samples whose predictions are in the core RAM vocabulary. \emph{Primary Score} is the \emph{Prediction Score} when rewarding only primary predictions. \emph{Overall Score} is the \emph{Prediction Score} with $\times$0.5 applied to the score contributions of any coarse predictions, effectively penalizing vague generic predictions. See \Cref{tab:tagging_model_comparison} for Wiki-H results. ${}^\dagger$Using DFN-5B H/14-378 CLIP model.}
    \label{tab:tagging_model_comparison_world}
\end{table*}

Training on large-scale datasets is also inherently not deterministic---in particular also due to the randomized data (\eg noise) augmentation---and some natural variations in the results will occur even if taking the mean of three training runs. The object decoder has thousands to tens of thousands of concepts to learn, and evaluating on a classification dataset with \eg 10--101 classes is probing a small fraction of what the model has learned and is rewarded for by its training loss. This means that two models with identical loss will always have some concepts that they have learned better, and some they have not learned quite as robustly. If that happens to align with the classes required for a classification dataset, the models then either do slightly better or slightly worse on them. As previously mentioned, CIFAR-10 is an example of all models doing about the same, up to natural variation, as the dataset classes are very few and coarse, and the dataset is relatively `easy'.

The classification results for the FT6 and FT2 NOVIC models are similar for many of the benchmarks, as both training sets have a reasonable size and comfortably cover all required class names as target nouns. The dip in performance for FT0 is more pronounced in general, due to the near-quadrupling relative to FT2 of the number of object concepts to learn, with the magnitude of the dip being tendentially correlated to larger numbers of classes and more fine-grained classification concepts overall. Both of these factors result in smaller regions of the embedding space, with less cosine distance between them, needing to be kept apart. This, in combination with the more finely `concept-subdivided' embedding space of an FT0 model, leads to the performance dip.

\section{Additional Baseline Comparisons}
\label{app:additional_results}

In \Cref{tab:tagging_model_comparison,tab:tagging_model_comparison_world}, we compare (as described in \Cref{sec:zero_shot_perf}) NOVIC to the image tagging baselines Tag2Text \cite{huang_tag2text_2024} and RAM \cite{zhang_recognize_2024}, on the Wiki-H and World-H datasets respectively. While the image tagging baselines require a list of noun candidates to be supplied at inference time and are trained on image-text pairs, NOVIC performs completely prompt-free image classification and is trained much more efficiently on text-only data. Also, while NOVIC is trained primarily on a synthetic text dataset constructed in a controllable and unbiased manner from an entire English dictionary of object nouns, the baselines train on `only' 14M image-text pairs, which have a significantly lower diversity of seen objects, and do not offer any controllability of the distributions of the nouns and image embeddings seen during training. Furthermore, Tag2Text has a fixed set of labels, while RAM attempts to provide some level of open vocabulary recognition beyond its nominal set of labels by distilling an image encoder from the base CLIP image encoder. Due to the limited size of the used image-text dataset however, and noting that the core set of labels was already \emph{constructed} by parsing these exact text captions, one can understand that the `open' abilities of RAM are somewhat limited, \ie RAM is not truly open vocabulary in any way close to how one considers CLIP to exhibit open vocabulary abilities.

In \Cref{tab:tagging_model_comparison,tab:tagging_model_comparison_world}, we can see that NOVIC has seen significantly more diverse object nouns during training (\numprint{42919} as opposed to less than a few thousand for the baselines), and this is clearly reflected by its ability to use a wide array of object nouns in its predictions---generally close to as many unique nouns as there are images in the corresponding dataset (\eg 263 unique predictions for 272 images). The tendency of the baselines not to provide such a diverse set of predictions can be observed from their significantly lower number of used objects. The restricted ability of RAM to perform open vocabulary classifications can also be identified by observing the high proportion of its generated predictions that belong to the core RAM vocabulary, even when provided with a large variety of object nouns to choose from (\eg FT0). RAM's open vocabulary predictions are also observed to have a significantly higher error rate than core vocabulary predictions.

Three prediction scores are calculated for each model---\emph{Prediction Score}, the standard prediction score as defined in \Cref{sec:experiments}, \emph{Primary Score}, which does not reward secondary predictions and thereby emphasizes the important ability of the model to focus on the actual intended object of an image,\footnote{Note that even the primary score does not adequately penalize the baselines' scores in practice however, for example in instances where a plate of food is the focus of the image and the baselines invariably predict \emph{plate} whereas NOVIC tends to predict the specific name of the dish.} and \emph{Overall Score}, which adjusts the standard prediction score to reduce the reward for overly coarse predictions and thereby emphasizes the important ability of the model to provide meaningful and informative fine-grained classifications. It should be noted that both the primary and overall scores can by definition only at most be equal to the standard prediction score, so an ideal model that emits only \emph{fine-grained primary} predictions would have all three scores be equal. We can observe that for NOVIC this is indeed almost the case, whereas the baselines exhibit very significant drops in score when penalizing secondary and coarse predictions as such. Overall, this demonstrates that NOVIC produces high quality predictions that are simultaneously diverse and fine-grained yet at the same time accurate, as well as correctly focusing on the main object of an image, all while being real-time, prompt-free, truly open vocabulary, and very efficiently trained with text only. This makes NOVIC a strong improvement over the state-of-the-art baselines.

\section{Ethical AI}
\label{app:ethical_ai}

Our open vocabulary image classifier may inherit biases from the CLIP model it is trained on, and be similarly susceptible to adversarial attacks during inference, such as manipulated images misguiding label generation. Although the object decoder is generative, it is trained in a constrained setting, making it unlikely to produce sensitive outputs. Training the object decoder on the proposed text datasets is significantly more cost-effective than training alternative methods such as RAM on large-scale image datasets. However, due to the large amount of noise used, identifying the optimal model configuration often necessitates many training runs, each with an environmental impact.

\end{document}